\documentclass[10pt,twocolumn,letterpaper]{article}
\usepackage[accsupp]{axessibility}  % camera-ready 
\usepackage[pagenumbers]{cvpr} % To force page numbers, e.g. for an arXiv version
\usepackage{adjustbox}
\usepackage{multirow}
\definecolor{cvprblue}{rgb}{0.21,0.49,0.74}
\usepackage[pagebackref,breaklinks,colorlinks,allcolors=cvprblue]{hyperref}

\usepackage{fancyhdr} % 引入自定义页脚宏包

\fancypagestyle{firstpage}{
    \fancyhf{}
    
    \fancyfoot[C]{\small % 用小一号字更美观
        Accepted for publication in the IEEE/CVF Conference on Computer Vision and Pattern Recognition (CVPR), 2026. \\
        \textcopyright\ 2026 IEEE. Personal use permitted; other uses require IEEE permission.
    }
}

\usepackage{bbm}
\usepackage{multirow}

\def\paperID{14125} % *** Enter the CVPR Paper ID here
\def\confName{CVPR}
\def\confYear{2026}
\title{Cross-Slice Knowledge Transfer via Masked Multi-Modal Heterogeneous Graph Contrastive Learning for Spatial Gene Expression Inference}

\author{
\hbox{Zhiceng Shi$^{1}$, Changmiao Wang$^{2}$, Jun Wan$^{3}$, Wenwen Min$^{1}$\textsuperscript{*}} 
\\
$^1$Yunnan University, Kunming 650500, China\\
$^2$Shenzhen Research Institute of Big Data, Shenzhen 518172, China\\
$^3$Zhongnan University of Economics and Law, Wuhan 430073, China\\
{\tt \small minwenwen@ynu.edu.cn}
}

\begin{document}

\maketitle
% 强制首页使用自定义的 fancy 页脚（article 类默认首页是 plain 样式，需单独指定）
% \thispagestyle{fancy}
\thispagestyle{firstpage} 

\begingroup
\renewcommand\thefootnote{\fnsymbol{footnote}}
\footnotetext[1]{Corresponding author.}
% \footnotetext[2]{Accepted for publication in the IEEE/CVF Conference on Computer Vision and Pattern Recognition (CVPR), 2026. © 2026 IEEE. }
\endgroup

\begin{abstract}
While spatial transcriptomics (ST) has advanced our understanding of gene expression in tissue context, its high experimental cost limits its large-scale application. Predicting ST from pathology images is a promising, cost-effective alternative, but existing methods struggle to capture complex cross-slide spatial relationships. 
To address the challenge, we propose SpaHGC, a multi-modal heterogeneous graph-based model that captures both intra-slice and inter-slice spot-spot relationships from histology images. 
It integrates local spatial context within the target slide and cross-slide similarities computed from image embeddings extracted by a pathology foundation model. 
These embeddings enable inter-slice knowledge transfer, and SpaHGC further incorporates Masked Graph Contrastive Learning to enhance feature representation and transfer spatial gene expression knowledge from reference to target slides, enabling it to model complex spatial dependencies and significantly improve prediction accuracy.
We conducted comprehensive benchmarking on seven matched histology-ST datasets from different platforms, tissues, and cancer subtypes. The results demonstrate that SpaHGC significantly outperforms the existing nine state-of-the-art methods across all evaluation metrics.  Additionally, the predictions are significantly enriched in multiple cancer-related pathways, thereby highlighting its strong biological relevance and application potential. Code and data are available at \href{https://github.com/wenwenmin/SpaHGC}{https://github.com/wenwenmin/SpaHGC}.
\end{abstract}

\begin{figure}[ht]
\centering
\includegraphics[width=1\columnwidth]{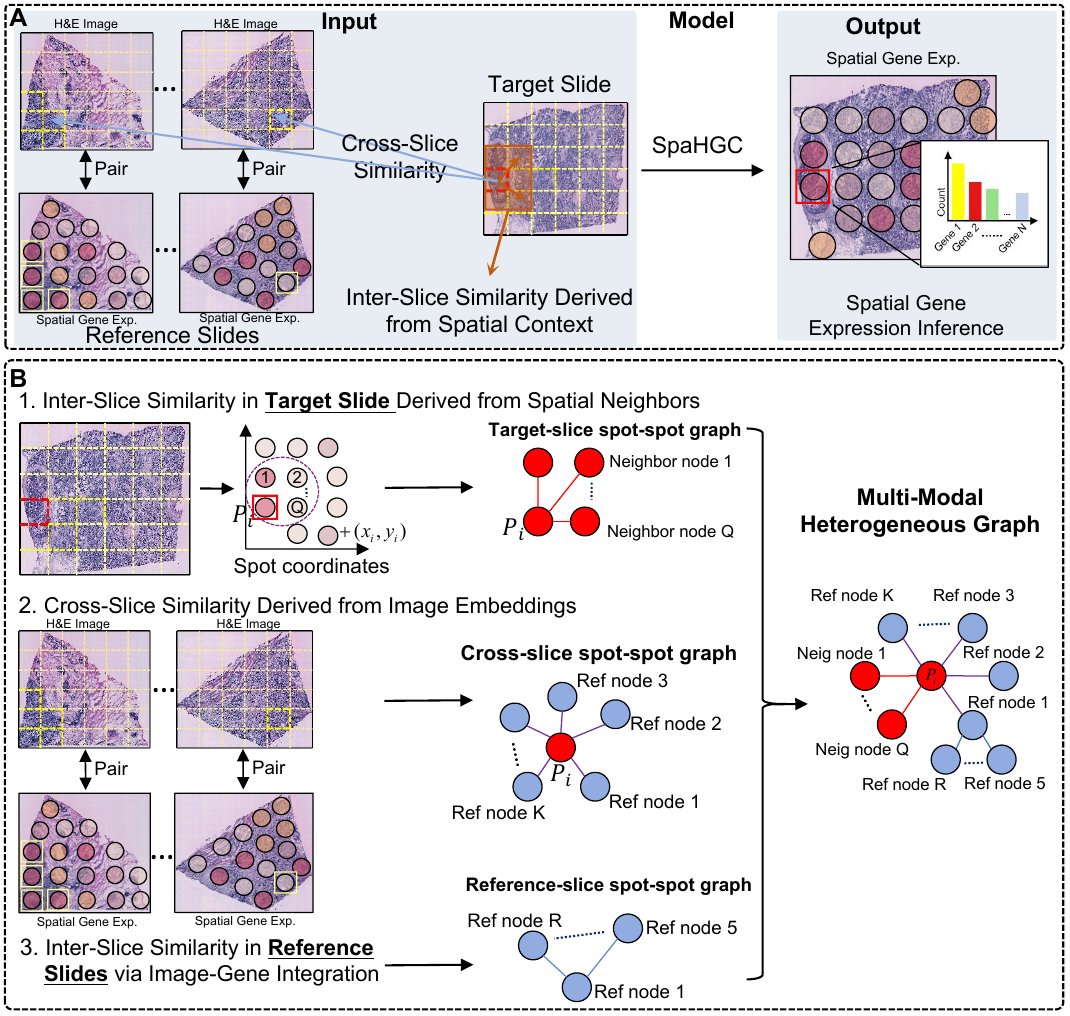}
\caption{
Overview of SpaHGC. 
(\textbf{A}) The inputs, outputs, and objectives of SpaHGC. 
%(\textbf{B}) The process of constructing the multi-modal heterogeneous graph from spatial transcriptomics and histology data.
(\textbf{B}) Key Contribution of SpaHGC: Constructing a multi-modal heterogeneous graph from cross-slice spatial transcriptomics and histology data.
}
\label{fig1}
\end{figure}

%%%%%%%%%%%%%%%%%%%%%%%%%%%%%
\section{Introduction}
The rapid development of Spatial Transcriptomics (ST) technology has enabled precise quantification of gene expression along with its spatial localization within tissue sections, providing a novel perspective for studying gene function \cite{song2025spatially,gandin2025deep,oh2025global}. In typical ST protocols, gene expression is profiled at hundreds to thousands of spatially localized spots across tissue sections. However, the high cost and technical complexity of ST protocols remain major barriers to their widespread adoption in large-scale research and clinical settings \cite{tian2023, STINR}. In contrast, matched hematoxylin and eosin (H\&E)-stained pathology images are routinely acquired in clinical workflows, being both inexpensive and widely available \cite{HisToGene,xue2026spacrd,xue2026inferring}. Consequently, researchers have begun exploring low-cost alternatives for predicting ST gene expression profiles directly from pathology images. Notably, each ST spot can be paired with a corresponding image patch in the pathology image, forming a natural spatial alignment between transcriptomic and morphological data \cite{jaume2024transcriptomics,zhao2025spaidl}. This correspondence provides a critical structural foundation for predicting spatial gene expression using pathology images.

Some studies have focused on directly predicting spatial gene expression from pathology images. These approaches \cite{STNet, HisToGene, His2ST, THItogene, HGGEP} typically use the image-spot paired data generated by ST as supervisory signals. By training an image encoder to extract features from pathology images, the models perform regression tasks to predict the gene expression profiles of the corresponding spots. In addition to this direct prediction strategy, some studies \cite{Bleep, mclSTExp} have introduced multimodal contrastive learning frameworks to align pathological image features with gene expression features in a shared latent space. Other works \cite{TRIPLEX, M2OST} have explored the integration of multi-scale image features to mimic the diagnostic process of pathologists, enhancing the model’s ability to capture both local and global tissue architecture. Furthermore, prior knowledge~\cite{EGGN, M2TGLGO, MERGE, huang2025stpath} has been leveraged to enhance prediction performance by integrating structural or biological constraints.

Although these existing methods have made progress in ST data analysis, most studies have not fully addressed the inherent characteristics of ST data. On the one hand, ST data often exhibit sparsity and noise—gene expression may be missing or extremely low at certain spatial locations, and data quality can be affected by experimental resolution, sample preparation, and batch effects \cite{SPACEL, Spavae, spamask, PRAGA, fang2026adapting}. On the other hand, current approaches mainly focus on modeling spatial structures within a single tissue slice, overlooking potentially shared expression patterns across different slides. In reality, individuals with the same tissue type or disease often share morphological and spatial gene expression similarities \cite{kueckelhaus2024inferring,EGGN, M2TGLGO, ohglobal}, while individual variability and disease progression introduce heterogeneity across samples. This diversity poses challenges for models to learn generalizable representations from single slides. Therefore, effectively integrating shared information across slides while accounting for individual-specific differences is crucial for improving expression prediction and enhancing model generalization.

To address these challenges, we propose SpaHGC, a novel masked multi-modal heterogeneous graph contrastive learning framework designed to jointly model intra- and inter-slice spot relationships, aiming to enhance the accuracy of spatial gene expression prediction (Figure \ref{fig1}). SpaHGC constructs a multi-modal heterogeneous graph with three distinct subgraphs: a Target-slice spot-spot graph to capture local spatial semantics, a Cross-slice spot-spot graph connecting target spots with reference spots using concatenated image and gene expression features to enable knowledge transfer, and a Reference-slice spot-spot graph within each reference slice to capture global expression relationships (Figure \ref{fig1}A and B). To capture shared expression patterns and suppress inter-slide differences, we introduce the Cross Node Dual Attention (CNDA) and Cross Node Attention Pooling (CNAP) modules. The two modules selectively update features across different node types and dynamically aggregate relevant reference node information, enabling cross-slide knowledge transfer. Furthermore, to enhance robustness, SpaHGC employs node-type-specific feature masking to simulate dropout and sequencing noise.

Our main contributions can be summarized as follows:
\begin{itemize}
\item We propose SpaHGC, a novel masked multi-modal heterogeneous graph learning framework that simultaneously models local spatial structures within individual slices and shared expression patterns across slices, effectively enhancing the accuracy and generalizability of gene expression prediction.

\item We design a complementary node masking strategy and introduce the CNDA and CNAP modules. These strategies and modules enable robust and efficient cross-slice knowledge transfer.

\item SpaHGC achieves state-of-the-art results on seven ST datasets across diverse platforms, tissues, and cancer subtypes, with improvements ranging from 7.3\% to 27.1\% in Pearson Correlation Coefficient (PCC) metrics over nine baseline methods. Subsequent biological downstream analyses further demonstrate the model’s practical utility.
\end{itemize}

\begin{figure*}[t]
\centering
\includegraphics[width=1\textwidth]{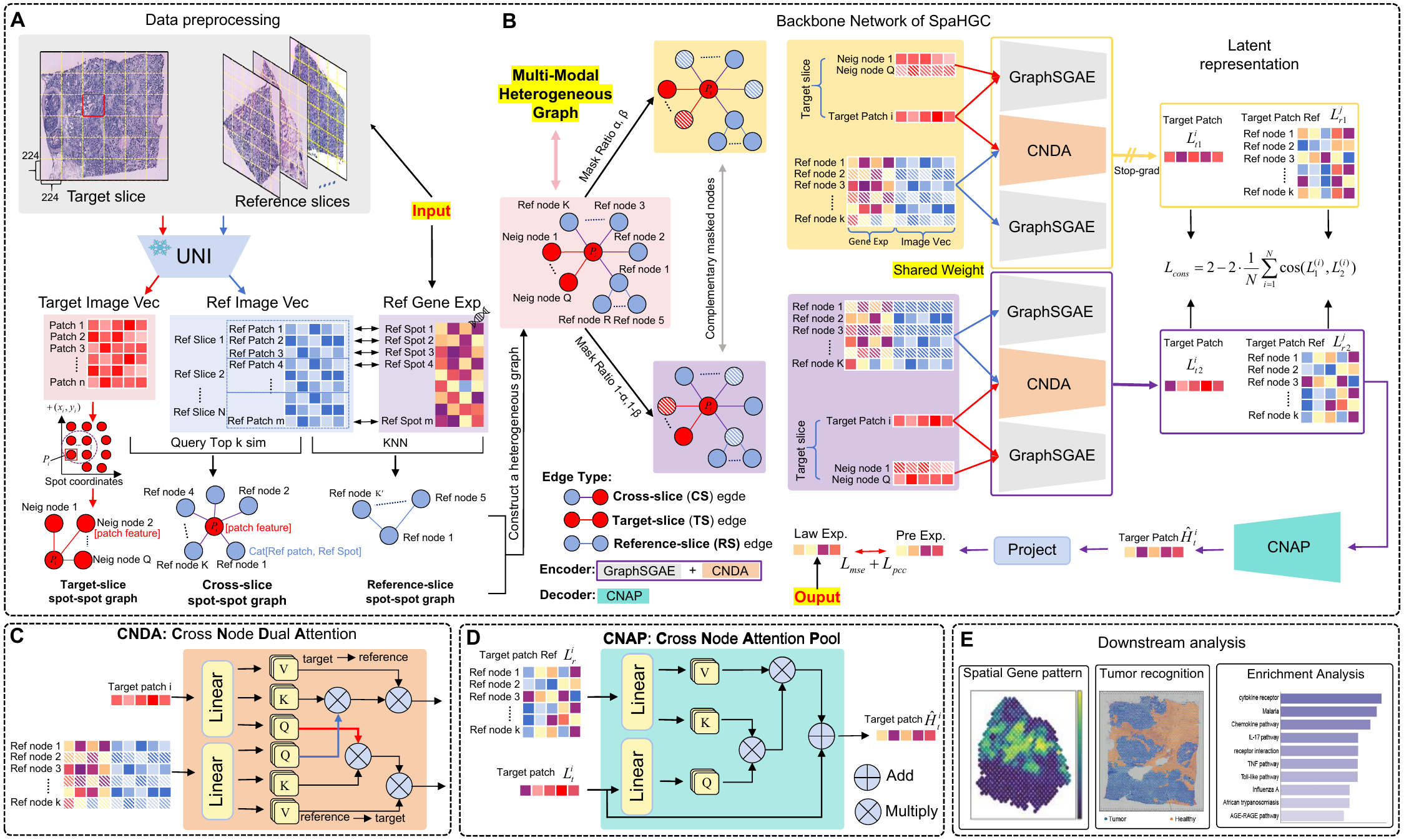}
\caption{
The overall architecture of SpaHGC. 
(\textbf{A}) Data preprocessing and construction of the multi-modal heterogeneous graph. 
(\textbf{B}) The backbone network of SpaHGC. 
(\textbf{C}) Architecture of the CNDA module. 
(\textbf{D}) Architecture of the CNAP module. 
(\textbf{E}) Downstream analysis.
}
\label{fig2}
\end{figure*}

\section{Related Work}
Existing approaches for spatial expression prediction from histology images fall into three main categories.

\textbf{Image-to-expression mapping.} Early approaches mainly focused on directly mapping  pathology images to spatial gene expression profiles through supervised regression models. Representative works such as STNet \cite{STNet} train convolutional neural networks (CNNs) to predict the expression of individual genes from local tissue patches. While these methods are computationally efficient and intuitive, they primarily capture local morphological features and ignore spatial context or relationships between neighboring spots.

\textbf{Intra-Slide Spatial Modeling.} To overcome the locality limitation of pure image-to-expression mappings, intra-slide spatial modeling methods have been developed to capture spatial dependencies among neighboring spots within the same tissue section. For instance, HisToGene \cite{HisToGene}, Hist2ST \cite{His2ST}, and THItoGene \cite{THItogene} incorporate spot-level spatial adjacency into graph-based architectures, enabling message passing between neighboring spots. HGGEP \cite{HGGEP} and M2OST \cite{M2OST} further enhance spatial modeling by integrating multi-scale contextual features from pathology images, mimicking the hierarchical reasoning process of pathologists. These intra-slide approaches demonstrate substantial performance improvements over simple CNN baselines, as they exploit the geometric structure of ST data. However, they remain constrained to single-slide analysis and cannot effectively leverage cross-slice similarities or shared patterns across different tissue samples.

\textbf{Multimodal alignment.} A complementary line of work aims to align pathology images and gene-expression profiles within a shared latent space, enabling expression prediction through cross-modal retrieval rather than direct regression. BLEEP \cite{Bleep} jointly encodes H\&E patches and expression vectors using a CLIP-style \cite{clip} contrastive objective and infers expression by retrieving and aggregating the nearest reference profiles. mclSTExp \cite{mclSTExp} similarly learns a joint embedding by integrating spatial context through a Transformer encoder and aligning patch–spot pairs via contrastive learning, allowing expression estimation through nearest-neighbor aggregation. EGGN \cite{EGGN} further incorporates exemplar retrieval into a graph framework, linking windows to matched exemplars and propagating molecular information through message passing.

However, although these methods provide complementary strengths in local structural modeling and multimodal embedding, they generally fail to effectively integrate information across slides and lack global context, which limits their ability to capture complex spatial expression patterns.

\section{Methodology}
\subsection{Problem Formulation}
Our objective is to predict spatial gene expression profiles from histopathological images. Formally, each ST sample is denoted as a tuple \((\mathbf{X}_t, \mathbf{y}_t)\), where \(\mathbf{X}_t \in \mathbb{R}^{3 \times H \times W}\) is a tissue image and \(\mathbf{y}_t \in \mathbb{R}^{n \times M}\) is the expression matrix for \(M\) genes across \(n\) spatial spots. Given a target sample \((\mathbf{X}_t, \mathbf{y}_t)\), the goal is to accurately estimate its expression matrix \(\hat{\mathbf{y}}_t\) by leveraging the visual information in \(\mathbf{X}_t\) along with auxiliary information derived from a set of reference slices. To this end, we construct a multi-modal heterogeneous graph \(\mathbf{G} = (\mathbf{V}, \mathbf{E})\) (Figure \ref{fig1}B). Based on the constructed \(\mathbf{G}\), we train a heterogeneous graph neural network to predict gene expression profiles at each spatial location on the target slide.

\subsection{Heterogeneous Graph Construction}
As illustrated in Figure~\ref{fig2}A, each slide is first partitioned into image patches centered on the ST spot coordinates. Let $P_t \in \mathbb{R}^{n \times 3 \times 224 \times 224}$ and $P_r \in \mathbb{R}^{m \times 3 \times 224 \times 224}$ denote the patch sets from the target and reference slides respectively, where $n$ and $m$ are the numbers of spots. We extract high-dimensional visual embeddings using a pretrained pathology foundation model UNI \cite{UNI}: 
\begin{equation}
    \mathbf{z}_t = \text{UNI}(P_t) \in \mathbb{R}^{n \times d}, \quad 
    \mathbf{z}_r = \text{UNI}(P_r) \in \mathbb{R}^{m \times d}
\end{equation}
where $d=1024$ is the embedding dimension.

We define two types of nodes in the heterogeneous graph. Each target node \(v_t^{(i)} \in V_t\) is associated with a patch embedding \(\mathbf{z}_t^{(i)}\), while each reference node \(v_r^{(j)} \in V_r\) is associated with a concatenated vector \(\mathbf{h}^{(j)}_{r} = [\mathbf{z}_r^{(j)} \, \| \, \mathbf{y}_r^{(j)}] \in \mathbb{R}^{d+M}\), combining visual and gene expression features.

\begin{table*}[htp]
\caption{Comparison with baseline methods on seven public ST datasets spanning different platforms, tissue types, and cancer subtypes.}
\centering
\begin{adjustbox}{width=1\textwidth}
\fontsize{16}{22}\selectfont 
\begin{tabular}{l|cccccccccccccccccccc}
\hline
\multicolumn{1}{l|}{\multirow{2}{*}{\textbf{Methods}}} & \multicolumn{2}{c}{HER+} &  & \multicolumn{2}{c}{cSCC} &  & \multicolumn{2}{c}{Alex} &  & \multicolumn{2}{c}{Visium BC} &  & \multicolumn{2}{c}{Lymph Node} &  & \multicolumn{2}{c}{Pancreas1} &  & \multicolumn{2}{c}{Pancreas2} \\ \cline{2-3} \cline{5-6} \cline{8-9} \cline{11-12} \cline{14-15} \cline{17-18} \cline{20-21} 
\multicolumn{1}{c|}{} & PCC(\%) $\uparrow$ & RMSE $\downarrow$ &  & PCC(\%) $\uparrow$ & RMSE $\downarrow$  &  & PCC(\%) $\uparrow$ & RMSE $\downarrow$ &  & PCC(\%) $\uparrow$ & RMSE $\downarrow$ &  & PCC(\%) $\uparrow$ & RMSE $\downarrow$ &  & PCC(\%) $\uparrow$ & RMSE $\downarrow$ &  & PCC(\%) $\uparrow$ & RMSE $\downarrow$ \\ \hline
\multirow{2}{*}{STNet \cite{STNet}} & 5.61 & 0.16 &  & 9.2 & 0.23 &  & 3.2 & 0.29 &  & 2.8 & 0.26 &  & 3.4 & 0.27 &  & 2.24 & 0.28 &  & 31.56 & 0.26 \\
 & $\pm$0.016 & $\pm$0.003 &  & $\pm$0.011 & $\pm$0.002 &  & $\pm$0.034 & $\pm$0.003 &  & $\pm$0.005 & $\pm$0.003 &  & $\pm$0.017 & $\pm$0.002 &  & $\pm$0.027 & $\pm$0.001 &  & $\pm$0.013 & $\pm$0.003 \\
\multirow{2}{*}{HisToGene \cite{HisToGene}} & 7.89 & 0.14 &  & 17.56 & 0.22 &  & 1.11 & 0.24 &  & 1.16 & 0.21 &  & 19.24 & 0.25 &  & 1.65 & 0.26 &  & 26.13 & 0.29 \\
 & $\pm$0.034 & $\pm$0.002 &  & $\pm$0.018 & $\pm$0.003 &  & $\pm$0.002 & $\pm$0.002 &  & $\pm$0.006 & $\pm$0.001 &  & $\pm$0.004 & $\pm$0.004 &  & $\pm$0.006 & $\pm$0.003 &  & $\pm$0.008 & $\pm$0.002 \\
\multirow{2}{*}{His2ST \cite{His2ST}} & 14.43 & 0.15 &  & 19.23 & 0.19 &  & 11.94 & 0.23 &  & 14.63 & 0.28 &  & 9.34 & 0.26 &  & 2.33 & 0.22 &  & 13.67 & 0.26 \\
 & $\pm$0.027 & $\pm$0.005 &  & $\pm$0.027 & $\pm$0.001 &  & $\pm$0.016 & $\pm$0.001 &  & $\pm$0.009 & $\pm$0.007 &  & $\pm$0.004 & $\pm$0.003 &  & $\pm$0.009 & $\pm$0.002 &  & $\pm$0.004 & $\pm$0.004 \\
\multirow{2}{*}{EGGN \cite{EGGN}} & 17.98 & 0.15 &  & 16.22 & 0.17 &  & 3.25 & 0.2 &  & 12.26 & 0.21 &  & 24.78 & 0.24 &  & 13.81 & 0.21 &  & 30.1 & 0.25 \\
 & $\pm$0.034 & $\pm$0.001 &  & $\pm$0.012 & $\pm$0.005 &  & $\pm$0.002 & $\pm$0.001 &  & $\pm$0.004 & $\pm$0.003 &  & $\pm$0.003 & $\pm$0.001 &  & $\pm$0.008 & $\pm$0.004 &  & $\pm$0.005 & $\pm$0.002 \\
\multirow{2}{*}{THItoGene \cite{THItogene}} & 17.26 & 0.16 &  & 18.15 & 0.21 &  & 10.69 & 0.23 &  & 2.34 & 0.21 &  & 10.82 & 0.27 &  & 2.16 & 0.27 &  & 18.97 & 0.27 \\
 & $\pm$0.011 & $\pm$0.006 &  & $\pm$0.001 & $\pm$0.002 &  & $\pm$0.005 & $\pm$0.006 &  & $\pm$0.002 & $\pm$0.009 &  & $\pm$0.001 & $\pm$0.001 &  & $\pm$0.011 & $\pm$0.005 &  & $\pm$0.003 & $\pm$0.003 \\
\multirow{2}{*}{HGGEP \cite{HGGEP}} & 19.68 & 0.13 &  & 20.13 & 0.23 &  & 2.55 & 0.26 &  & 7.79 & 0.22 &  & 6.2 & 0.23 &  & 2.3 & 0.24 &  & 5.65 & 0.25 \\
 & $\pm$0.022 & $\pm$0.005 &  & $\pm$0.001 & $\pm$0.003 &  & $\pm$0.004 & $\pm$0.006 &  & $\pm$0.006 & $\pm$0.002 &  & $\pm$0.009 & $\pm$0.002 &  & $\pm$0.005 & $\pm$0.003 &  & $\pm$0.006 & $\pm$0.004 \\
\multirow{2}{*}{BLEEP \cite{Bleep}} & 18.56 & 0.18 &  & 23.56 & 0.24 &  & 5.04 & 0.23 &  & 4.09 & 0.24 &  & 24.83 & 0.23 &  & 2.29 & 0.22 &  & 28.55 & 0.26 \\
 & $\pm$0.012 & $\pm$0.002 &  & $\pm$0.022 & $\pm$0.002 &  & $\pm$0.021 & $\pm$0.003 &  & $\pm$0.011 & $\pm$0.002 &  & $\pm$0.006 & $\pm$0.002 &  & $\pm$0.008 & $\pm$0.001 &  & $\pm$0.005 & $\pm$0.005 \\
\multirow{2}{*}{mclSTExp \cite{mclSTExp}} & 23.15 & 0.13 &  & 31.88 & 0.26 &  & 6.77 & 0.21 &  & 5.13 & 0.22 &  & 21.64 & 0.25 &  & 9.22 & 0.25 &  & 31.61 & 0.22 \\
 & $\pm$0.021 & $\pm$0.003 &  & $\pm$0.007 & $\pm$0.005 &  & $\pm$0.023 & $\pm$0.002 &  & $\pm$0.003 & $\pm$0.002 &  & $\pm$0.019 & $\pm$0.003 &  & $\pm$0.051 & $\pm$0.002 &  & $\pm$0.011 & $\pm$0.004 \\
\multirow{2}{*}{M2OST \cite{M2OST}} & 18.24 & 0.14 &  & 24.88 & 0.25 &  & 15.13 & 0.22 &  & 6.52 & 0.21 &  & 30.97 & 0.24 &  & 15.12 & 0.26 &  & 38.35 & 0.23 \\
 & $\pm$0.017 & $\pm$0.002 &  & $\pm$0.005 & $\pm$0.002 &  & $\pm$0.026 & $\pm$0.0023 &  & $\pm$0.016 & $\pm$0.003 &  & $\pm$0.015 & $\pm$0.002 &  & $\pm$0.017 & $\pm$0.002 &  & $\pm$0.021 & $\pm$0.003 \\
\multirow{2}{*}{SpaHGC (Ours)} & \textbf{27.86} & \textbf{0.12} & \textbf{} & \textbf{38.79} & \textbf{0.15} & \textbf{} & \textbf{17.19} & \textbf{0.19} & \textbf{} & \textbf{20.08} & \textbf{0.19} & \textbf{} & \textbf{35.02} & \textbf{0.23} & \textbf{} & \textbf{24.48} & \textbf{0.17} & \textbf{} & \textbf{41.36} & \textbf{0.21} \\
 & \textbf{$\pm$0.028} & \textbf{$\pm$0.001} & \textbf{} & \textbf{$\pm$0.003} & \textbf{$\pm$0.0014} & \textbf{} & \textbf{$\pm$0.004} & \textbf{$\pm$0.006} & \textbf{} & \textbf{0.004} & \textbf{$\pm$0.001} & \textbf{} & \textbf{$\pm$0.003} & \textbf{$\pm$0.004} & \textbf{} & \textbf{$\pm$0.005} & \textbf{$\pm$0.003} & \textbf{} & \textbf{$\pm$0.003} & \textbf{$\pm$0.006} \\ \hline
\end{tabular}
\end{adjustbox}
\label{tab1}
\end{table*}
Adjacent spots are typically located within similar morphological regions and exhibit correlated gene expression patterns. To capture spatial continuity within the target tissue, we connect neighboring spots based on their Euclidean distances. Given the spot coordinates
$\mathbf{c}_t = \{\mathbf{s}_t^{(i)} \in \mathbb{R}^2\}_{i=1}^{n}$, we define the Target-slice (TS) edge set to capture local spatial semantics:
\begin{equation}
    d_{ij} = \left\| \mathbf{s}_t^{(i)} - \mathbf{s}_t^{(j)} \right\|_2
\end{equation}

\begin{equation}
    \mathcal{A}_i = \operatorname{Top}Q_j(-d_{ij}), \quad |\mathcal{A}_i| = Q
\end{equation}

\begin{equation}
    E_{\text{TS}} = \bigcup_{i=1}^{n} \left\{ (v_t^{(i)}, v_t^{(j)}) \;\middle|\; j \in \mathcal{A}_i \right\}
\end{equation}
where \(\operatorname{Top}Q_j(-d_{ij})\) selects the \(Q\) closest neighbors.

Because a single slide's tissue morphology cannot fully represent complex gene expression patterns, we incorporate multiple reference slides of the same tissue type to establish cross-slide connections. Specifically, for each target patch embedding \(\mathbf{z}_{t}^{(i)}\), we query its Top-\(K\) most similar reference patch embeddings from \(\mathbf{z}_r \in \mathbb{R}^{m \times 1024}\)  based on cosine similarity:
\begin{equation}
\mathcal{N}_i = \text{TopK}\left( \left\{ \cos\left( \mathbf{z}_{t}^{(i)}, \mathbf{z}_{r}^{(j)} \right) \mid j = 1, \dots, m , \right\} \right),
\end{equation}
where \(\mathcal{N}_i\) is the set of Top-\(K\) reference patches most similar to the \(i\)-th target patch, and $\left | \mathcal{N}_i \right | =K$.
We then define the Cross-slice (CS) edges set connecting target nodes to their corresponding reference nodes:
\begin{equation}
    E_{\text{CS}} = \bigcup_{i=1}^{n} \left\{ (v_t^{(i)}, v_r^{(j)}) \;\middle|\; j \in \mathcal{N}_i \right\}.
\end{equation}

To exploit the expression and morphological patterns within the reference cohort, we further construct intra-reference connections. Specifically, we introduce a set of Reference-slice (RS) edges \(E_{\text{RS}}\) that link reference nodes with their most similar counterparts based on joint features. These connections allow reference nodes to share information across the entire cohort, reinforcing their representations with globally consistent semantic and expression context. To implement this, for each reference patch embedding \(\mathbf{z}_r^{(j)}\), we identify its Top-\(K\) most similar peers based on cosine similarity:
\begin{equation}
\scalebox{0.85}{$
\mathcal{C}_j = \text{Top}K\left( \left\{ \cos\left( \mathbf{h}_r^{(j)}, \mathbf{h}_r^{(k)} \right) \mid k \neq j,\; k = 1, \dots, m \right\} \right),
$}
\end{equation}

Based on these similarity neighborhoods, the RS edge set is defined as:
\begin{equation}
E_{\text{RS}} = \bigcup_{j=1}^{m} \left\{ \left(v_r^{(j)}, v_r^{(k)}\right) \mid k \in \mathcal{C}_j \right\}.
\end{equation}

To summarize, the constructed heterogeneous graph integrates multi-level spatial and semantic relationships across both the target and reference slides. For each target node \(v^{(i)}_{t}\) with visual embedding \(z^{(i)}_{t}\), TS edges \(E_{\text{TS}}\) connect \(v^{(i)}_{t}\) to to its $Q$ nearest neighbors within the same target slide and capture local spatial continuity and tissue context. CS edges \(E_{\text{CS}}\) link \(v^{(i)}_{t}\) to its Top-$K$ most similar reference nodes \(v^{(j)}_{r}\) enabling cross-slide knowledge transfer based on morphological similarity. RS edges \(E_{\text{RS}}\) further connect each reference node \(v^{(j)}_{r}\) to its Top-$K$ most similar reference peers \(v^{(k)}_{r}\), facilitating information exchange within the reference cohort and serving as a global semantic scaffold for expression representation. 

\begin{figure*}[htp]
\centering
\includegraphics[width=1\textwidth]{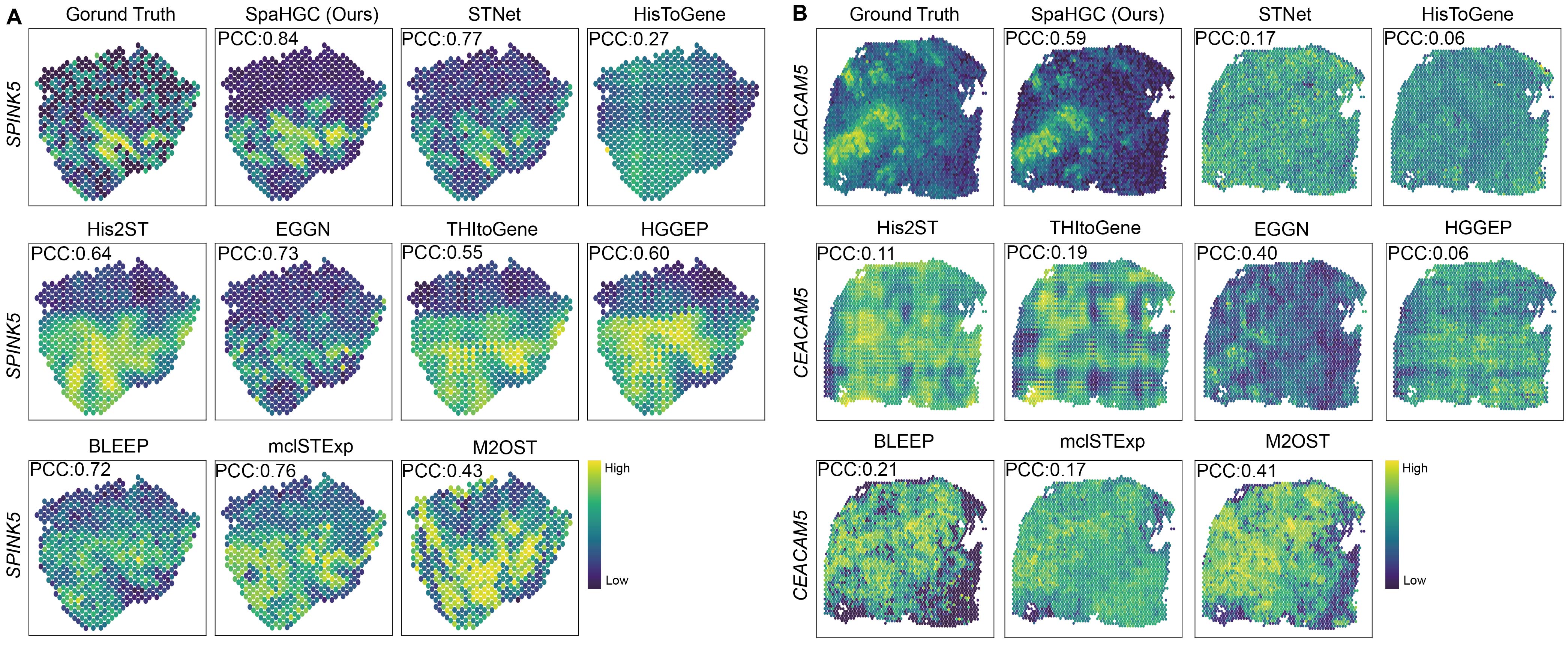}
\caption{Visualization of predicted expression patterns for marker genes. Results are shown for SpaHGC and nine baseline methods. (\textbf{A}) \textit{SPINK5} on the cSCC dataset. (\textbf{B}) \textit{CEACAM5} on the Pancreas1 dataset.}
\label{fig3}
\end{figure*}

\subsection{Complementary Masking Strategy}
To enhance the model’s robustness against inherent noise in ST data, we propose a complementary masking strategy (Figure \ref{fig2}B). This approach generates two semantically distinct yet topologically identical augmented graph views, simulating potential feature loss and perturbations present in real data. Consequently, the model is encouraged to learn stable and consistent representations despite noise interference. Concretely, we apply node-type-specific feature masking separately to the neighbor nodes of target nodes and to the reference nodes. For each view \(k \in \{1,2\}\), we construct binary masks \(\mathbf{M}_t^{(k)} \in \{0,1\}^{n \times d}\) and \(\mathbf{M}_r^{(k)} \in \{0,1\}^{m \times (d+M)}\), with masking ratios $\alpha$ and $\beta$, respectively, which control the fraction of features that are masked out. The two views are constructed to be feature-wise complementary, satisfying:
\begin{equation}
\mathbf{M}_t^{(1)} + \mathbf{M}_t^{(2)} = \mathbf{1}, \quad 
\mathbf{M}_r^{(1)} + \mathbf{M}_r^{(2)} = \mathbf{1}
\end{equation}

\begin{equation}
\tilde{\mathbf{z}}_t^{(k)} = \mathbf{z}_t \odot \mathbf{M}_t^{(k)}, \quad 
\tilde{\mathbf{h}}_r^{(k)} = \mathbf{h}_r \odot \mathbf{M}_r^{(k)}
\end{equation}
where \(\mathbf{1}\) denotes an all-one matrix. The corresponding augmented graphs:
\[
\tilde{G}^{(k)} = (V, E, \tilde{\mathbf{Z}}^{(k)}), \quad \tilde{\mathbf{Z}}^{(k)} = \tilde{\mathbf{z}}_t^{(k)} \cup \tilde{\mathbf{h}}_r^{(k)}.
\]

To encode the augmented views \(\tilde{G}^{(k)}\),\(k \in \{1,2\}\), we construct a heterogeneous graph encoder with shared parameters across the two views. The encoder comprises three distinct modules, each designed for a specific type of relational edge: two GraphSAGE \cite{hamilton2017inductive} modules for modeling \(E_{Ts}\) and \(E_{Rs}\) relations, and a CNDA module for capturing \(E_{CS}\) interactions between target and reference nodes:

\begin{equation}
\begin{aligned}
L^{(k)}_{t} &= \text{GraphSAGE}(V_{t}, E_{\mathrm{TS}}, \tilde{\mathbf{z}}^{(k)}_{t}) \\
\bar{L}^{(k)}_{\mathrm{t}},  \bar{L}^{(k)}_{\mathrm{r}}&= \text{CNDA}(V_{t} \cup V_{r}, E_{\mathrm{CS}}, \tilde{\mathbf{z}}^{(k)}_{t} \cup \tilde{\mathbf{h}}^{(k)}_{r}) \\
L^{(k)}_{r} &= \text{GraphSAGE}(V_{r}, E_{\mathrm{RS}}, \tilde{\mathbf{h}}^{(k)}_{r})
\end{aligned}
\end{equation}

We update the representations of both node types by averaging their local and cross-attentional embeddings using the scatter mean operation typical.

\begin{equation}
\begin{aligned}
\hat{L}^{(k)}_{t} &= \text{scatter\_mean}(L^{(k)}_{t}, \bar{L}^{(k)}_{t}) \\
\hat{L}^{(k)}_{r} &= \text{scatter\_mean}(L^{(k)}_{r}, \bar{L}^{(k)}_{r}) \\
\end{aligned}
\end{equation}

Finally, we introduce a self-supervised contrastive learning objective that encourages the embeddings of the same target node from two views to be close in the semantic space. We adopt an asymmetric contrastive design, where one view is processed through the encoder without gradient updates, serving as a stable target, while the other view is passed through the same encoder with standard backpropagation enabled. The two sets of node embeddings are then compared using a contrastive loss to encourage consistency across views. Specifically, we minimize the cosine distance between the corresponding node embeddings from the two views:
\begin{equation}\label{eq:14}
\scalebox{0.9}{$
    \mathcal{L}_{\text{con}}^{i} = \frac{1}{N} \sum_{j=1}^{N} \left( 2 - 2 \cdot \text{Cos}\left( \hat{L}^{1}_{j}, \hat{L}^{2}_{j} \right) \right), \quad i \in \{t, r\}
    $}
\end{equation}
where $N$ denotes the total number of nodes participating in the contrastive loss, including both target and reference nodes.

\subsection{Cross Node Dual Attention (CNDA) Module}
To capture shared expression patterns across slides and suppress inter-slide variability for effective cross-slide knowledge transfer, CNDA employs a bidirectional attention mechanism that establishes dynamic communication channels between target and reference nodes (Figure \ref{fig2}C).

Specifically, each target node performs cross-node dual attention with its corresponding reference nodes along the referential edge set \(E_{CS}\). Notably, the feature vector of each reference node is defined as \(\mathbf{h}^{(j)}_{r} = [\mathbf{z}_r^{(j)} \, \| \, \mathbf{y}_r^{(j)}]\), which combines visual and gene expression features. Through dynamically computed attention weights, the target node can selectively transfer embedded morphological and gene expression knowledge from the reference nodes: 
\begin{equation}
\mathbf{Q}_t = \mathbf{z}_t \mathbf{W}_Q^t, \quad
\mathbf{K}_r = \mathbf{h}_r \mathbf{W}_K^r, \quad
\mathbf{V}_r = \mathbf{h}_r \mathbf{W}_V^r
\end{equation}
\begin{equation}
\mathbf{A}_{t \leftarrow r} = \mathrm{softmax}\left( \frac{\mathbf{Q}_t \mathbf{K}_r^\top}{\sqrt{d'}} \right)
\end{equation}
\begin{equation}
\bar{\mathbf{L}}_t = \mathbf{A}_{t \leftarrow r} \mathbf{V}_r
\end{equation}
where, \(\mathbf{W}_Q^t, \mathbf{W}_K^r, \mathbf{W}_V^r \in \mathbb{R}^{d \times {d}' }  \) are learnable weight matrices. Symmetrically, reference nodes attend to the visual embeddings of target nodes to update their own features:
\begin{equation}
\mathbf{Q}_r = \mathbf{h}_r \mathbf{W}_Q^r, \quad
\mathbf{K}_t = \mathbf{z}_t \mathbf{W}_K^t, \quad
\mathbf{V}_t = \mathbf{z}_t \mathbf{W}_V^t
\end{equation}
\begin{equation}
\mathbf{A}_{r \leftarrow t} = \mathrm{softmax}\left( \frac{\mathbf{Q}_r \mathbf{K}_t^\top}{\sqrt{d'}} \right)
\end{equation}
\begin{equation}
\bar{\mathbf{L}}_r = \mathbf{A}_{r \leftarrow t} \mathbf{V}_t
\end{equation}
where, \(\mathbf{W}_Q^r, \mathbf{W}_K^t, \mathbf{W}_V^t \in \mathbb{R}^{d \times {d}' }  \) are learnable weight matrices. \(\bar{\mathbf{L}}_t\) semantically summarizes the existing visual and transcriptomic knowledge, while \(\bar{\mathbf{L}}_r\) encodes the expected gene expression knowledge. 

\subsection{Cross Node Attention Pool (CNAP) Module}
To enhance the feature representation capability of target nodes, we propose the CrossNodeAttentionPool (CNAP) module, which adaptively integrates auxiliary information from reference nodes based on the contextual semantics of each target node (Figure \ref{fig2}D). Compared to existing integration strategies \cite{EGGN}, we introduce a multi-head unidirectional cross-node attention mechanism, computing cross-attention between the target representation \(\hat{\mathbf{L}}_{t}\) and the reference representation \(\hat{\mathbf{L}}_{r}\), thereby guiding the information fusion process. Formally, the CNAP block is defined as:
\begin{equation}
\scalebox{0.85}{$
\mathbf{Q}_t^{(i)} = \hat{\mathbf{L}}_t \mathbf{W}_Q^{t,(i)}, \quad
\mathbf{K}_r^{(i)} = \hat{\mathbf{L}}_r \mathbf{W}_K^{r,(i)}, \quad
\mathbf{V}_r^{(i)} = \hat{\mathbf{L}}_r \mathbf{W}_V^{r,(i)},
$}
\end{equation}
\begin{equation}
\scalebox{0.85}{$
\mathbf{A}^{(i)} = \mathrm{softmax}\left(\frac{\mathbf{Q}_t^{(i)} (\mathbf{K}_r^{(i)})^\top}{\sqrt{d'/h}}\right), \quad
\mathbf{H}_t^{(i)} = \mathbf{A}^{(i)} \mathbf{V}_r^{(i)}
$}
\end{equation}
\begin{equation}
\hat{\mathbf{H}}_t = \mathrm{Concat}\left(\mathbf{H}_t^{(1)}, \ldots, \mathbf{H}_t^{(h)}\right) \mathbf{W}_O
\end{equation}
where \(h\) is the number of attention heads, \(i \in \left \{ 1,\dots,h \right \} \) indexes the attention heads, \(\mathbf{W}_Q^{t,(h)}, \mathbf{W}_K^{r,(h)}, \mathbf{W}_V^{r,(h)} \in \mathbb{R}^{d \times \frac{d'}{H}}\) are the learnable projection matrices for the \(h\)-th head, and \(\mathbf{W}_O \in \mathbb{R}^{d' \times d'}\) is the output projection matrix.

Finally, to more effectively model the contribution of reference nodes to the gene expression prediction of target nodes and to guide the model to integrate features with greater reliance on the target nodes' own representations, we introduce a residual connection during the fusion stage:
\begin{equation}
\hat{\textbf{y}}_{t} = \text{MLP}\left (  \hat{\mathbf{L}}_{t} + \hat{\mathbf{H}}_{t} \right ) 
\end{equation}

\subsection{Loss Function}
The model’s optimization objective includes minimizing the mean squared error (MSE) loss, maximizing the Pearson correlation coefficient (PCC) loss, and minimizing the contrastive loss between two augmented views:
\begin{equation}
\mathcal{L}_{mse} = \frac{1}{N} \sum_{i=1}^{N} \left \|   \textbf{y}_{t} - \hat{\textbf{y}}_{t}  \right \| ^2_2 
\end{equation}
\begin{equation}
\mathcal{L}_{pcc}=1-\frac{\mathrm{Cov}(\textbf{y}_{t}, \hat{\textbf{y}}_{t})}{\mathrm{Var}(\textbf{y}_{t}) \cdot \mathrm{Var}(\hat{\textbf{y}}_{t})}
\end{equation}

\begin{equation}
\mathcal{L}_{total}= \mathcal{L}_{mse} + \mathcal{L}_{pcc} + \mathcal{L}^{t}_{con} + \mathcal{L}^{r}_{con}
\end{equation}
where \(\mathcal{L}^{t}_{con}\) is the contrastive loss between augmented views of target nodes, and 
\(\mathcal{L}^{r}_{con}\) is the contrastive loss between augmented views of reference nodes (See Eq \ref{eq:14}).

\section{Experiments}
% To comprehensively evaluate SpaHGC’s performance, we compared it with nine state-of-the-art methods across seven matched histology-ST datasets covering various platforms, tissue types, and cancer subtypes (Supplementary Table S1). Detailed information on datasets, data preprocessing, evaluation metrics, and model efficiency assessment can be found in \textbf{Supplementary Material} S2.
% \subsection{Implementation Details}

\subsection{Experimental Setup}

\paragraph{Datasets.}

We evaluated SpaHGC on seven publicly available matched image–ST datasets spanning a variety of cancer types and tissues as listed in Supplementary Table S1. These datasets consist of cutaneous squamous cell carcinoma profiled using the ST platform, three breast cancer datasets including a \textit{HER2+} subtype profiled via the ST platform, the Alex dataset representing the \textit{TNBC} subtype on the Visium platform and the Visium BC dataset of the \textit{HER2+} subtype also on the Visium platform, one lymph node dataset generated with the Visium platform as well as two pancreatic cancer datasets Pancreas1 and Pancreas2 both acquired using the Visium platform. Further details are provided in Section \ref{AppendDatasets} of the Supplementary Materials.

\paragraph{Baselines and Evaluation Metrics.}
To evaluate the performance of SpaHGC, we compared it with nine state-of-the-art methods, including one image-to-expression mapping: STNet \cite{STNet}, five intra-slide spatial modeling: HisToGene \cite{HisToGene}, Hist2ST \cite{His2ST}, THItoGene \cite{THItogene}, HGGEP \cite{HGGEP}, M2OST \cite{M2OST}, and multimodal alignment: EGGN \cite{EGGN}, Bleep \cite{Bleep} and mclSTExp \cite{mclSTExp}. Sections \ref{AppendBaseline} of the Supplementary Materials provide detailed descriptions of the baselines. 

We use Pearson Correlation Coefficient (PCC) and Root Mean Square Error (RMSE) to evaluate the predictive performance of the model. Mathematically, PCC and RMSE can be described as:
\begin{equation}
\mathrm{PCC} = \frac{\mathrm{Cov}(\textbf{y}_{t}, \hat{\textbf{y}}_{t})}{\mathrm{Var}(\textbf{y}_{t}) \cdot \mathrm{Var}(\hat{\textbf{y}}_{t})}
\end{equation}
\begin{equation}
\mathrm{RMSE} = \sqrt{ \frac{1}{N} \sum_{i=1}^N (\textbf{y}_t - \hat{\textbf{y}}_t)^2 }
\end{equation}
where $\textbf{y}_{t}$ denotes the ground truth values, $\hat{\textbf{y}}_{t}$represents the predicted values, $Cov(\cdot, \cdot)$ is the covariance, $Var(\cdot)$ denotes the variance.

In addition, we employ the Adjusted Rand Index (ARI) to evaluate the accuracy of the clustering results:
\begin{equation}
\text{ARI}= \frac{ {\textstyle \sum_{ij}\binom{n_{ij}}{2}}-\frac{\left [ {\textstyle \sum_{i}\binom{a_{i}}{2}\sum_{j}\binom{b_{j}}{2}}\right] }{\binom{n}{2}}}{\frac{1}{2} \left  [\sum_{i}\binom{a_{i}}{2}+\sum_{j}\binom{b_{j}}{2}\right ]- \frac{\left [ {\textstyle \sum_{i}\binom{a_{i}}{2}\sum_{j}\binom{b_{j}}{2}}\right] }{\binom{n}{2}}}, 
\end{equation}
where $a_{i}$ and $b_{j}$ represent the number of samples assigned to the $i$-th predicted cluster and the $j$-th ground truth cluster, respectively, while $n_{ij}$ denotes the number of samples that are shared between the $i$-th predicted cluster and the $j$-th ground truth cluster.

\paragraph{Implementation Details.}
We conducted all experimentsusing a single NVIDIA RTX 4090 GPU, with the development environment based on PyTorch 2.1.2 and Python 3.9. The SpaHGC model consists of 4 layers of GraphSAGE and CNDA modules, along with one CNAP module employing 4-head multi-head attention. The model was trained for 200 epochs on all datasets, with a learning rate of $1 \times 10^{-4}$ and weight decay of $1 \times 10^{-4}$. The number of neighbor nodes was set to 5 ($Q$), and the number of reference nodes was set to 7  ($K$). In the augmented views, the feature masking ratios were set to 0.8 ($\alpha$) for target nodes and 0.9 ($\beta$) for reference nodes, respectively. All baseline methods used the default hyperparameters reported in their respective papers. All models were trained on a Windows operating system with Python 3.9 and PyTorch 2.1.2, using an NVIDIA RTX 4090 GPU with 24GB of memory.

\paragraph{Preventing Data Leakage in Nested LOOCV.} 

To ensure rigorous evaluation and eliminate potential data leakage, we adopt a nested leave-one-out cross-validation (Nested LOOCV) strategy, illustrated using the cSCC dataset as an example. In the outer loop, one slide is held out as the test set, while the remaining slides are used for training and validation. Within each outer training set, an inner LOOCV is conducted for hyperparameter tuning using only the training data, ensuring that the outer test sample is never involved in model selection or optimization. To further prevent leakage during reference retrieval, the Top-K reference neighbors for each target slide are drawn exclusively from the corresponding fold’s training reference bank, strictly excluding the target slide itself and all validation/test slides.

\subsection{Experimental Results}
\paragraph{SpaHGC enables accurate inference of spatial gene expression profiles.}
As shown in Table \ref{tab1}, SpaHGC consistently achieves the highest PCC and lowest MSE across all datasets, with PCC improvements ranging from 7.3\% to 27.1\%, underscoring its superiority in spatial gene expression prediction. Supplementary Figures \ref{figs1}–\ref{figs3} further demonstrate that SpaHGC outperforms other methods on each individual slide. Paired Wilcoxon signed-rank tests yielded p-values $<$ 0.001 across all datasets (Supplementary Table \ref{tableS2}), confirming the statistical significance of the performance gains.

\paragraph{SpaHGC can reconstruct gene expression patterns.}

Numerical metrics alone do not fully reflect a model’s predictive performance. Gene expression patterns, as key indicators of tissue function, contain crucial spatial regulatory information. Thus, accurately reconstructing these patterns is essential for evaluating biological interpretability and practical utility.

Figure \ref{fig3} shows the visualization of expression patterns for the marker genes \textit{SPINK5} in the cSCC dataset and \textit{CEACAM5} in the Pancreas1 dataset. Compared to baseline methods, SpaHGC better preserves original expression patterns, demonstrating superior spatial representation. Furthermore, Supplementary Figures \ref{figs4}–\ref{figs8} further confirm SpaHGC’s consistent robustness in reconstructing gene expression patterns across all datasets.

\paragraph{SpaHGC exhibits strong performance in tumor region localization.}
The gene expression profiles of spots in ST data reflect tissue functional states and pathological heterogeneity, serving as a crucial basis for accurate tumor region identification. To evaluate the tumor region identification ability of various methods, we performed clustering based on predicted expression on the Alex and VisiumBC datasets, using pathologist-annotated tissue regions as the gold standard to classify spots into tumor and non-tumor areas. The results (Supplementary Figures \ref{figS9}-\ref{figs10}) show that SpaHGC achieves the highest ARI, significantly outperforming other methods and demonstrating superior accuracy in tumor region identification.

\paragraph{SpaHGC yields gene expression predictions with functional biological relevance.}

To assess the biological relevance of gene expression predicted by SpaHGC, we performed differential expression analysis on the Alex dataset to identify genes significantly differing between spots labeled as “invasive cancer” and “non-tumor tissue.” These genes were then subjected to Kyoto Encyclopedia of Genes and Genomes (KEGG) and Gene Ontology Biological Process (GO:BP) pathway enrichment analyses.

As shown in Supplementary Figure \ref{figs11}, several canonical pathways closely associated with tumor initiation and progression were identified, such as \textit{Viral protein interaction with cytokine and cytokine receptor} \cite{spangler2015insights} and \textit{B Cell Receptor Signaling Pathway} \cite{BCR}. Compared to the results based on the original expression data, the enrichment results derived from the predicted expression exhibited greater statistical significance, suggesting that the expression predicted by SpaHGC is more closely aligned with the biological characteristics of cancerous tissue.

\subsection{Ablation Study}
In this section, we conduct systematic ablation studies on the Lymph Node, Pancreas1 and Pancreas2 datasets. We also compare computational time and model complexity as detailed in Section S6 of the Supplementary Materials.

\paragraph{Ablation study on the graph structure.}To systematically evaluate the role of different edge types in the model, we conducted ablation studies based on the three types of edges (\(E_{TS}, E_{CS}, E_{RS}\)). Specifically, we individually removed each type of edge to block its corresponding information flow, thereby independently assessing the contribution of each edge type to feature representation and overall model performance. As shown in Table \ref{tab2}, the removal of different edge types led to decreases in PCC by 1.3\%-6.9\%, 3.5\%-10.3\%, and 3.1\%-6.4\% on the three datasets, respectively. Notably, the removal of \(E_{CS}\) edges led to the most significant performance drop, indicating that this type of edge plays a crucial role in modeling cross-slice knowledge transfer.

%%%%%%%%%%%%%%%%%%%%%%%%%%%%%%%%%
\begin{table}[htp]
\centering
\caption{Ablation study on the graph structure.}
\begin{adjustbox}{width=0.47\textwidth}
\fontsize{15}{25}\selectfont
\begin{tabular}{l|lclllllll}
\hline
\multirow{2}{*}{Graph Structure} &  & \multicolumn{2}{c}{Lymph Node} &  & \multicolumn{2}{c}{Pancreas1} &  & \multicolumn{2}{c}{Pancreas2} \\
&  & PCC(\%) $\uparrow$ & \multicolumn{1}{c}{RMSE $\downarrow$} &  & \multicolumn{1}{c}{PCC(\%) $\uparrow$} & \multicolumn{1}{c}{RMSE $\downarrow$} &  & \multicolumn{1}{c}{PCC(\%) $\uparrow$} & \multicolumn{1}{c}{RMSE $\downarrow$} \\ \cline{1-4} \cline{6-7} \cline{9-10} 
w/o \(E_{TS}\) &  & 33.68$\pm$0.005 & \multicolumn{1}{c}{0.237$\pm$0.002} &  & \multicolumn{1}{c}{22.68$\pm$0.003} & 0.194$\pm$0.003 &  & 39.86$\pm$0.002 & 0.228$\pm$0.004 \\
w/o \(E_{CS}\) &  & 32.58$\pm$0.004 & 0.241$\pm$0.004 &  & 21.94$\pm$0.002 & 0.197$\pm$0.002 &  & 39.05$\pm$0.003 & 0.235$\pm$0.006 \\
w/o \(E_{RS}\) &  & 34.58$\pm$0.002 & 0.228$\pm$0.003 &  & 23.62$\pm$0.004 & \textbf{0.175$\pm$0.002} &  & 40.45$\pm$0.001 & 0.221$\pm$0.005 \\
\textbf{Ours} & \textbf{} & \textbf{35.02$\pm$0.003} & \textbf{0.225$\pm$0.004} & \textbf{} & \textbf{24.48$\pm$0.005} & 0.179$\pm$0.003 & \textbf{} & \textbf{41.76$\pm$0.003} & \textbf{0.213$\pm$0.006} \\ \hline
\end{tabular}
\end{adjustbox}
\label{tab2}
\end{table}
%%%%%%%%%%%%%%%%%%%%%%%%%%%%%%%%%
\begin{table}[htp]
\centering
\caption{Ablation study on the Image Encoder.}
\begin{adjustbox}{width=0.47\textwidth}
\fontsize{15}{25}\selectfont 
\begin{tabular}{l|lclllllll}
\hline
\multirow{2}{*}{Image Encoder} &  & \multicolumn{2}{c}{Lymph Node} &  & \multicolumn{2}{c}{Pancreas1} &  & \multicolumn{2}{c}{Pancreas2} \\
 &  & PCC(\%) $\uparrow$ & \multicolumn{1}{c}{RMSE $\downarrow$} &  & \multicolumn{1}{c}{PCC(\%) $\uparrow$} & \multicolumn{1}{c}{RMSE $\downarrow$} &  & \multicolumn{1}{c}{PCC(\%) $\uparrow$} & \multicolumn{1}{c}{RMSE $\downarrow$} \\ \cline{1-4} \cline{6-7} \cline{9-10} 
w/ Resnet18 \cite{resnet} &  & 32.57$\pm$0.004 & \multicolumn{1}{c}{0.233$\pm$0.001} &  & \multicolumn{1}{c}{20.25$\pm$0.005} & 0.193$\pm$0.001 &  & 38.94$\pm$0.001 & 0.218$\pm$0.003 \\
w/ Desnet121 \cite{Densnet} &  & 33.23$\pm$0.002 & 0.230$\pm$0.001 &  & 21.60$\pm$0.003 & 0.192$\pm$0.004 &  & 39.21$\pm$0.002 & 0.222$\pm$0.005 \\
w/ ViT-Base \cite{ViT} &  & 33.77$\pm$0.004 & 0.229$\pm$0.004 &  & 21.62$\pm$0.004 & 0.185$\pm$0.002 &  & 39.01$\pm$0.001 & 0.226$\pm$0.003 \\
w/ DeiT-Base \cite{DeiT} &  & 32.86$\pm$0.002 & 0.234$\pm$0.003 &  & 21.37$\pm$0.003 & 0.189$\pm$0.003 &  & 39.47$\pm$0.001 & 0.219$\pm$0.002 \\
w/ Virchow \cite{virchow} &  & 34.04$\pm$0.002 & 0.231$\pm$0.002 &  & 22.46$\pm$0.004 & 0.182$\pm$0.001 &  & 40.79$\pm$0.002 & 0.217$\pm$0.004 \\
\textbf{w/ UNI \cite{UNI} (Ours)} & \textbf{} & \textbf{35.02$\pm$0.003} & \textbf{0.225$\pm$0.004} & \textbf{} & \textbf{24.48$\pm$0.005} & \textbf{0.179$\pm$0.003} & \textbf{} & \textbf{41.76$\pm$0.003} & \textbf{0.213$\pm$0.006} \\ \hline
\end{tabular}
\end{adjustbox}

\label{tab3}
\end{table}
%%%%%%%%%%%%%%%%%%%%%%%%%%%%%%%%%
\begin{table}[htp]
\centering
\caption{Ablation study of the complementary masking strategy.}
\begin{adjustbox}{width=0.47\textwidth}
\fontsize{15}{25}\selectfont
\begin{tabular}{l|lcllcllll}
\hline
 \multirow{2}{*}{Masking Strategy} &  & \multicolumn{2}{c}{Lymph Node} &  & \multicolumn{2}{c}{Pancreas1} &  & \multicolumn{2}{c}{Pancreas2} \\
 &  & PCC(\%) $\uparrow$ & \multicolumn{1}{c}{RMSE $\downarrow$} &  & PCC(\%) $\uparrow$ & \multicolumn{1}{c}{RMSE $\downarrow$} &  & \multicolumn{1}{c}{PCC(\%) $\uparrow$} & \multicolumn{1}{c}{RMSE $\downarrow$} \\ \cline{1-4} \cline{6-7} \cline{9-10} 
w/o Mask &  & 32.59$\pm$0.003 & \multicolumn{1}{c}{0.245$\pm$0.003} &  & 22.19$\pm$0.002 & 0.199$\pm$0.004 &  & 37.99$\pm$0.003 & 0.231$\pm$0.004 \\
W/ Random Mask &  & 34.11$\pm$0.002 & 0.235$\pm$0.001 &  & \multicolumn{1}{l}{23.22$\pm$0.003} & 0.184$\pm$0.002 &  & 39.55$\pm$0.001 & 0.226$\pm$0.003 \\
\textbf{Ours} & \textbf{} & \textbf{35.02$\pm$0.003} & \textbf{0.225$\pm$0.004} & \textbf{} & \multicolumn{1}{l}{\textbf{24.48$\pm$0.005}} & \textbf{0.179$\pm$0.003} & \textbf{} & \textbf{41.76$\pm$0.003} & \textbf{0.213$\pm$0.006} \\ \hline
\end{tabular}
\end{adjustbox}

\label{tab4}
\end{table}
%%%%%%%%%%%%%%%%%%%%%%%%%%%%%%%%%
\begin{table}[]
\centering
\caption{Ablation study on the modules of SpaHGC.}
\begin{adjustbox}{width=0.47\textwidth}
\fontsize{15}{25}\selectfont
\begin{tabular}{l|lclllllll}
\hline
\multirow{2}{*}{Module} &  & \multicolumn{2}{c}{Lymph Node} &  & \multicolumn{2}{c}{Pancreas1} &  & \multicolumn{2}{c}{Pancreas2} \\
 &  & PCC(\%) $\uparrow$ & \multicolumn{1}{c}{RMSE $\downarrow$} &  & \multicolumn{1}{c}{PCC(\%) $\uparrow$} & \multicolumn{1}{c}{RMSE $\downarrow$} &  & \multicolumn{1}{c}{PCC(\%) $\uparrow$} & \multicolumn{1}{c}{RMSE $\downarrow$} \\ \cline{1-4} \cline{6-7} \cline{9-10} 
w/o CNDA &  & 33.68$\pm$0.002 & 0.236$\pm$0.001 &  & 22.87$\pm$0.003 & 0.185$\pm$0.002 &  & 39.69$\pm$0.010 & 0.228$\pm$0.011 \\
w/o CNAP &  & 34.74$\pm$0.001 & 0.233$\pm$0.003 &  & 23.79$\pm$0.003 & 0.186$\pm$0.003 &  & 40.76$\pm$0.002 & 0.221$\pm$0.007 \\
\textbf{Ours} & \textbf{} & \textbf{35.02$\pm$0.003} & \textbf{0.225$\pm$0.004} & \textbf{} & \textbf{24.48$\pm$0.003} & \textbf{0.179$\pm$0.003} & \textbf{} & \textbf{41.76$\pm$0.003} & \textbf{0.213$\pm$0.006} \\ \hline
\end{tabular}
\end{adjustbox}

\label{tab5}
\end{table}
%%%%%%%%%%%%%%%%%%%%%%%%%%%%%%%%%

\paragraph{Ablation study on the Image Encoder.}
We conducted ablation studies on various mainstream image encoders, including CNN architectures (ResNet18, DenseNet121), Transformer architectures (ViT, DeiT), and pathology-specific pretrained models (Virchow, UNI). The results (Table \ref{tab3}) show that pathology-specific pretrained models outperform traditional CNNs and general-purpose Transformers in extracting tissue image features. Among them, UNI achieves the best performance across multiple datasets and evaluation metrics.

\paragraph{Ablation study of the complementary masking strategy.} To evaluate the effectiveness of the proposed complementary feature masking strategy, we conducted a series of ablation experiments comparing the following three configurations: (1) no masking, where all input node features are retained; (2) random masking, where node features are randomly masked in each view; and (3) complementary masking, where mutually exclusive subsets of features are masked across the two augmented views to achieve semantic complementarity. The experimental results (Table \ref{tab4}) show that introducing feature masking significantly improves overall model performance, which suggests that this strategy may help the model better accommodate the noise, missing values, and uncertainty that are commonly observed in ST data. Among the three configurations, the complementary masking strategy consistently achieves the best performance across various evaluation metrics, further demonstrating its effectiveness in facilitating the learning of robust and generalizable representations through complementary feature views.
\paragraph{Ablation study on the modules of SpaHGC.}
To validate the effectiveness of the proposed modules, we conducted ablation studies by individually removing the CNDA and CNAP modules (Table \ref{tab5}). The CNDA module introduces a bidirectional attention mechanism that enables dynamic information exchange between target and reference nodes. The CNAP module during the feature fusion stage to selectively regulate reference information, thereby accommodating structural and semantic heterogeneity across tissue slices. Experimental results show that removing either module leads to a significant decline in model performance on cross-slice gene expression prediction tasks, further confirming the critical roles these modules play in modeling cross-slice knowledge transfer.

\paragraph{Ablation study on hyperparameters.}We conducted systematic ablation studies on the number of neighbor nodes, the number of reference nodes, as well as the masking ratios applied to both neighbor and reference nodes. By progressively adjusting these key hyperparameters, we evaluated their impact on model performance and generalization capability (Supplementary Table S3-S6). Based on a comprehensive consideration of model effectiveness, we ultimately set the number of neighbor nodes to 5, the number of reference nodes to 7, and the masking ratios for neighbor and reference nodes to 0.8 and 0.9, respectively.

\subsection{Computational Time vs. Model Complexity}
We evaluated the training and inference times of various models on the cSCC, Visium BC, and Pancreas2 datasets. To ensure a fair comparison, all models were run under the same hardware environment (NVIDIA RTX 4090 GPU) using the default hyperparameter settings recommended in their respective publications. As shown in Supplementary Figure S12, SpaHGC achieves the fastest computational time across all datasets. In terms of model complexity, it ranks second in the number of trainable parameters, slightly higher than M2OST \cite{M2OST}.

\section{Conclusion}
In this paper, we propose SpaHGC, a novel masked multi-modal heterogeneous graph neural network framework for predicting spatial gene expression from histopathological images. SpaHGC constructs a multi-modal heterogeneous graph to jointly model intra-slice spatial dependencies and inter-slice expression similarities, enabling the integration of morphological and transcriptomic information across samples. To facilitate effective cross-slice knowledge transfer while preserving individual-specific expression patterns, we introduce the CNDA module and the CNAP module. Moreover, we design a node-type-specific feature masking strategy to enhance model robustness and accuracy under common challenges in ST, such as missing values, sparsity, and noise. SpaHGC outperforms nine state-of-the-art methods on seven public ST datasets across diverse platforms, tissues, and cancer subtypes, with downstream analyses confirming the biological relevance of its predicted gene expression.

\section*{Acknowledgments}
The work was supported in part by the National Natural Science Foundation of China (No. 62262069) and the Program of Yunnan Key Laboratory of Intelligent Systems and Computing (No. 202405AV340009). 

\section*{Appendices}
 
\appendix
\renewcommand{\thesection}{\Alph{section}}
\setcounter{figure}{0}
\setcounter{table}{0}
\renewcommand{\thefigure}{S\arabic{figure}}
\renewcommand{\thetable}{S\arabic{table}}

\section{Experimental Datasets}
\label{AppendDatasets}
In our experiments, we evaluated the performance of the proposed SpaHGC across seven publicly available matched image-ST datasets covering a variety of cancer types and tissue contexts (Table \ref{tableS1}). These include one human skin squamous cell carcinoma (cSCC) dataset \cite{cSCC} from the traditional ST platform with 12 samples; three breast cancer datasets, comprising the HER2+ subtype from the ST platform with 36 samples \cite{HER2+}, and two Visium platform datasets—Alex (subtype: TNBC) \cite{Alex} with 4 samples and Visium BC (subtype: HER2+) \cite{VisiumBC} with 3 samples. 
Additionally, we incorporated the HEST-1k \cite{HEST1K} dataset to further evaluate the generalizability of our model across diverse tissue types. We selected 4 lymph node samples and two pancreatic cancer datasets (Pancreas1 and Pancreas2), containing 4 and 3 samples respectively.

The traditional ST platform typically samples several hundred spatially resolved spots per tissue section, each measuring expression levels for approximately 20,000 genes. In contrast, the Visium platform offers enhanced spatial resolution and sampling density, enabling the capture of thousands of spots per section while maintaining comparable gene coverage per spot.

\begin{table}[htp]
\renewcommand{\thetable}{S\arabic{table}}
\caption{Summary of seven matched image-ST datasets used in this study.}
\begin{adjustbox}{width=0.47\textwidth}
\fontsize{10}{15}\selectfont 
\begin{tabular}{lllccc}
\hline
Dataset Name & Reference & Cancer & Subtype & Platform & Slice Num  \\ \hline
cSCC & \cite{cSCC} & Squamous Cell Carcinoma & - & ST & 12  \\
HER2+ & \cite{HER2+} & Breast Cancer & HER2+ & ST & 36  \\
Alex & \cite{Alex} & Breast Cancer & TNBC & Visium & 4  \\
Visium BC & \cite{VisiumBC} & Breast Cancer & HER2+ & Visium & 3  \\
Lymph Node & \cite{HEST1K} & Lymph Node & - & Visium & 4 \\
Pancreas1 & \cite{HEST1K} & Pancreatic Cancer & - & Visium & 3  \\
Pancreas2 & \cite{HEST1K} & Pancreatic Cancer & - & Visium & 4  \\ \hline
\end{tabular}
\end{adjustbox}
\label{tableS1}
\end{table}

\subsection{Data Preprocessing}
Following Li et al.~\cite{HisToGene}, we adopted the following preprocessing pipeline for gene selection. For each tissue slice, we first applied the Scanpy toolkit to identify the top 1{,}000 highly variable genes (HVGs) using the function \textit{scanpy.pp.highly\_variable\_genes} on log-normalized counts. To obtain a consistent feature space across all slices within the same dataset, we then computed the intersection of the HVG sets from all slices and restricted subsequent analyses to this shared HVG set. As a result, the size of the shared HVG set varied across datasets, ranging from approximately 200 to 800 genes.

Subsequently, for each spot in all datasets, we normalized the gene expression counts by dividing each gene’s raw count by the total expression within that spot, followed by multiplication with a scaling factor of 1,000,000. The normalized values are subsequently transformed using the natural logarithm in the form of \(\mathrm{log}(1+x)\),
where \(x\) denotes the normalized expression value. For image preprocessing, we extract \(224 \times 224\)  pixel patches from the pathology slides based on the centroid coordinates of each spot, and use a pretrained pathology image model to extract their visual feature representations.

\section{Baselines}
\label{AppendBaseline}
To evaluate the performance of SpaHGC, we compared it with nine state-of-the-art methods methods as follows:

\textbf{STNet} \cite{STNet} is one of the earliest methods to explore image-driven expression prediction. It utilizes DenseNet-121 \cite{Densnet} to extract features from image patches corresponding to each spot and employs fully connected layers to regress the expression levels of multiple target genes. 

\textbf{HisToGene} \cite{HisToGene} introduces the Vision Transformer (ViT) \cite{ViT}, leveraging self-attention to capture global dependencies between image patches and enhance spatial structure representation.

\textbf{Hist2ST} \cite{His2ST} crops an image patch around each sequenced spot and uses a Convmixer module to capture intra-patch 2D visual features. Spot coordinates are encoded and fused with the features, and a Transformer models global spatial dependencies across the slide. A GraphSAGE layer then aggregates local neighborhood information on a 4-nearest-neighbor graph, with LSTM-based layer aggregation to combine multi-layer representations. Gene expression is modeled via a ZINB likelihood with an additional MSE loss, and a data-augmentation-driven self-distillation strategy is employed to improve data efficiency.

\textbf{EGGN} \cite{EGGN} predicts spot-level gene expression from pathology images by coupling exemplar retrieval with graph neural inference. Each slide is partitioned into windows, which are embedded with a ResNet-18 backbone to define a retrieval space; for every window, its nearest exemplars—with known expression profiles—are retrieved from a reference bank. A graph is then constructed whose nodes include both windows and exemplars, with four edge types encoding window–window spatial proximity, exemplar–exemplar similarity, and the bidirectional links between windows and their matched exemplars. Over this graph, a GraphSAGE backbone performs message passing within the window and exemplar subgraphs, while a Graph Exemplar Bridging block explicitly injects exemplar knowledge into window features using the exemplars’ expression signals. Finally, an attention-based prediction head weights each window’s exemplars and aggregates them with the updated window representation to produce gene-expression vectors for all windows.

\textbf{THItoGene} \cite{THItogene} first partitions each pathology images slide into spot-centered image patches and uses dynamic convolution to adaptively extract fine-grained, multi-scale visual representations. It then introduces an Efficient-CapsNet with attention-based routing to capture the hierarchical spatial relationships of cellular morphology and tissue architecture, thereby strengthening the modeling of the linkage between image phenotypes and gene expression. Next, spot coordinate embeddings are fused with deep visual features, and a ViT is employed to model global, cross-region dependencies. Finally, a 4-nearest-neighbor graph is constructed based on Euclidean distances between spots, over which a GAT performs attention-based aggregation to explicitly learn the correspondence between spatial neighborhoods and gene expression, yielding spot-level expression predictions.

\textbf{HGGEP} \cite{HGGEP} is a hypergraph neural network for predicting gene expression from pathology images. It first partitions each whole-slide image into spot-centered patches and enhances morphological cues with a Gradient Enhancement Module based on difference convolution. A lightweight ShuffleNet-V2 backbone then extracts multi-stage features, which are refined at each stage by channel and spatial attention and further processed with a Vision Transformer to capture long-range, cross-spot dependencies. To model higher-order relationships across stages, a Hypergraph Association Module constructs hyperedges using both Euclidean distance and local positional metrics, enabling joint reasoning over groups of spots. The resulting hypergraph representations are concatenated with ViT features and passed to an LSTM to promote inter-stage information flow, and a final MLP outputs spot-level gene expression predictions.

\textbf{BLEEP} \cite{Bleep} formulates pathology images-to-expression prediction as bi-modal representation learning followed by retrieval-based imputation: paired H\&E patches and spot-level expression profiles are encoded by an image encoder and an expression encoder into a shared low-dimensional space using a CLIP-style \cite{clip} contrastive objective that is smoothed by incorporating intra-batch image–image and expression–expression similarities, which prevents near-duplicate spots from being pushed apart in training; at inference, each query patch is projected into the joint space and its k nearest reference expression profiles are retrieved and linearly combined to yield the predicted expression vector, thereby shifting the task from per-gene regression to joint profile interpolation while leveraging morphological–molecular alignment learned from the reference.

\textbf{mclSTExp} \cite{mclSTExp} predicts gene expression from pathology images by learning a joint embedding between image patches and spatially contextualized spot expression. The model first encodes each spot-centered H\&E patch with an image encoder and projects it into a latent space, while treating spots as “tokens” and the whole slide as a “sentence” so that a Transformer with self-attention produces semantically consistent spatial representations. It then aligns the two modalities, bringing matched patch–spot pairs closer and pushing mismatches apart in cosine space, thereby yielding a multimodal representation in which morphology and expression are co-registered. At inference time, a query patch is mapped into this shared space, its top-k nearest reference spots are retrieved, and their ground-truth expression profiles are aggregated with weights to produce the gene-expression prediction for the corresponding slide.

\textbf{M2OST} \cite{M2OST} formulates pathology images to expression prediction as a many-to-one regression problem in which multi-level pathology images patches that share a common target spot are processed jointly to produce a single gene-expression vector. The model employs a decoupled multi-scale encoder: deformable patch embedding emphasizes the spot’s central region while generating fine-grained in-spot tokens and coarser surrounding tokens; intra-level token mixing performs ViT-style self-attention within each scale; cross-level token mixing introduces fully connected cross-scale attention to exchange information across scales with different sequence lengths; and cross-level channel mixing applies squeeze–excitation to fuse channels in a length-agnostic manner. The [CLS] tokens from all levels are finally concatenated and fed to a linear head to output spot-level expressions, and the architecture naturally accommodates variable numbers of inputs while leveraging nearby inter-spot cues inherent in multi-resolution pathology images.

% \clearpage
% \newpage
% \clearpage
% \input{sec/X_suppl}
{
    \small
    \bibliographystyle{ieeenat_fullname}
    \bibliography{main}
}

\newpage

\begin{table*}[]
\renewcommand{\thetable}{S\arabic{table}}
\caption{Wilcoxon signed-rank test p-values comparing SpaHGC with competing methods across different datasets. Each p-value is computed based on paired comparisons of PCC scores across all tissue slices within the dataset.}
\begin{adjustbox}{width=1\textwidth}
\fontsize{9}{14}\selectfont 
\begin{tabular}{l|ccccccc}
\hline
P-value & \multicolumn{1}{l}{HER2+} & \multicolumn{1}{l}{cSCC} & \multicolumn{1}{l}{Alex} & \multicolumn{1}{l}{Visium BC} & Lymph Node & Pancreas1 & Pancreas2 \\ \hline
SpaHGC vs. STNet & \textless{}0.001 & \textless{}0.001 & \textless{}0.001 & \textless{}0.001 & \textless{}0.001 & \textless{}0.001 & \textless{}0.001 \\
SpaHGC vs. HisToGene & \textless{}0.001 & \textless{}0.001 & \textless{}0.001 & \textless{}0.001 & \textless{}0.001 & \textless{}0.001 & \textless{}0.001 \\
SpaHGC vs. His2ST & \textless{}0.001 & \textless{}0.001 & \textless{}0.001 & \textless{}0.001 & \textless{}0.001 & \textless{}0.001 & \textless{}0.001 \\
SpaHGC vs. EGGN & \textless{}0.001 & \textless{}0.001 & \textless{}0.001 & \textless{}0.001 & \textless{}0.001 & \textless{}0.001 & \textless{}0.001 \\
SpaHGC vs. THItoGene & \textless{}0.001 & \textless{}0.001 & \textless{}0.001 & \textless{}0.001 & \textless{}0.001 & \textless{}0.001 & \textless{}0.001 \\
SpaHGC vs. HGGEP & \textless{}0.001 & \textless{}0.001 & \textless{}0.001 & \textless{}0.001 & \textless{}0.001 & \textless{}0.001 & \textless{}0.001 \\
SpaHGC vs. Bleep & \textless{}0.001 & \textless{}0.001 & \textless{}0.001 & \textless{}0.001 & \textless{}0.001 & \textless{}0.001 & \textless{}0.001 \\
SpaHGC vs. mclSTExp & \textless{}0.001 & \textless{}0.001 & \textless{}0.001 & \textless{}0.001 & \textless{}0.001 & \textless{}0.001 & \textless{}0.001 \\
SpaHGC vs. M2OST & \textless{}0.001 & \textless{}0.001 & \textless{}0.001 & \textless{}0.001 & \textless{}0.001 & \textless{}0.001 & \textless{}0.001 \\ \hline
\end{tabular}
\end{adjustbox}
\label{tableS2}
\end{table*}

\begin{table*}[htp]
\renewcommand{\thetable}{S\arabic{table}}
\caption{Ablation study on the number of neighbor nodes 
$Q$.}
\begin{adjustbox}{width=1\textwidth}
\fontsize{12}{17}\selectfont 
\begin{tabular}{l|lllllllll}
\hline
\multirow{2}{*}{Q Value} &  & \multicolumn{2}{c}{Lymph Node} &  & \multicolumn{2}{c}{Pancreas1} &  & \multicolumn{2}{c}{Pancreas2} \\
 &  & \multicolumn{1}{c}{PCC} & \multicolumn{1}{c}{RMSE} &  & \multicolumn{1}{c}{PCC} & \multicolumn{1}{c}{RMSE} &  & \multicolumn{1}{c}{PCC} & \multicolumn{1}{c}{RMSE} \\ \cline{1-4} \cline{6-7} \cline{9-10} 
Q=1 &  & 33.81±0.002 & 0.244±0.004 &  & 22.44±0.001 & 0.185±0.001 &  & 36.67±0.003 & 0.237±0.002 \\
Q=2 &  & 34.56±0.001 & 0.235±0.008 &  & 22.94±0.005 & 0.185±0.002 &  & 40.41±0.002 & 0.225±0.013 \\
Q=3 &  & 34.89±0.001 & 0.239±0.001 &  & 23.62±0.003 & 0.176±0.001 &  & 40.99±0.002 & 0.219±0.005 \\
Q=4 &  & 34.97±0.003 & 0.227±0.002 &  & 23.51±0.002 & \textbf{0.174±0.001} &  & 41.21±0.001 & 0.219±0.003 \\
Q=5 & \textbf{} & \textbf{35.02±0.003} & 0.225±0.004 & \textbf{} & \textbf{24.48±0.005} & 0.179±0.003 & \textbf{} & \textbf{41.76±0.003} & \textbf{0.213±0.006} \\
Q=6 &  & 34.78±0.002 & 0.227±0.002 &  & 24.02±0.003 & 0.175±0.001 &  & 41.23±0.001 & 0.215±0.002 \\
Q=7 &  & 34.22±0.002 & 0.224±0.007 &  & 24.26±0.001 & 0.172±0.002 &  & 41.04±0.002 & 0.216±0.001 \\
Q=8 &  & 34.40±0.002 & 0.226±0.001 &  & 23.59±0.003 & 0.175±0.003 &  & 41.26±0.001 & 0.215±0.001 \\
Q=9 &  & 34.86±0.001 & \textbf{0.221±0.008} &  & 23.80±0.003 & 0.176±0.001 &  & 41.32±0.002 & 0.219±0.003 \\ \hline
\end{tabular}
\end{adjustbox}
\label{tableS3}
\end{table*}

\begin{table*}[htp]
\renewcommand{\thetable}{S\arabic{table}}
\caption{Ablation study on the number of reference nodes $K$.}
\begin{adjustbox}{width=1\textwidth}
\fontsize{12}{17}\selectfont 
\begin{tabular}{l|lllllllll}
\hline
\multirow{2}{*}{K Value} &  & \multicolumn{2}{c}{Lymph Node} &  & \multicolumn{2}{c}{Pancreas1} &  & \multicolumn{2}{c}{Pancreas2} \\
 &  & \multicolumn{1}{c}{PCC} & \multicolumn{1}{c}{RMSE} &  & \multicolumn{1}{c}{PCC} & \multicolumn{1}{c}{RMSE} &  & \multicolumn{1}{c}{PCC} & \multicolumn{1}{c}{RMSE} \\ \cline{1-4} \cline{6-7} \cline{9-10} 
K=1 &  & 33.76±0.002 & 0.235±0.002 &  & 23.49±0.001 & 0.180±0.018 &  & 41.19±0.001 & 0.214±0.002 \\
K=2 &  & 34.24±0.001 & 0.230±0.004 &  & 23.50±0.001 & 0.182±0.002 &  & 41.39±0.003 & 0.214±0.001 \\
K=3 &  & 34.33±0.002 & 0.231±0.004 &  & 23.09±0.002 & 0.184±0.001 &  & 41.54±0.001 & 0.213±0.001 \\
K=4 &  & 34.21±0.001 & 0.228±0.003 &  & 23.51±0.001 & 0.183±0.002 &  & 41.73±0.003 & 0.216±0.001 \\
K=5 & \textbf{} & 34.43±0.001 & \textbf{0.222±0.005} &  & 24.29±0.003 & 0.182±0.003 & \textbf{} & 41.76±0.002 & 0.212±0.009 \\
K=6 &  & 34.27±0.003 & 0.226±0.001 &  & 24.15±0.003 & 0.185±0.002 &  & 41.75±0.004 & 0.219±0.007 \\
K=7 &  & \textbf{35.02±0.003} & 0.225±0.004 & \textbf{} & \textbf{24.48±0.005} & \textbf{0.179±0.003} & \textbf{} & \textbf{41.76±0.003} & \textbf{0.213±0.006} \\
K=8 &  & 34.49±0.002 & 0.228±0.001 &  & 24.05±0.002 & 0.186±0.001 &  & 41.61±0.001 & 0.213±0.011 \\
K=9 &  & 34.86±0.001 & 0.226±0.008 &  & 24.30±0.004 & 0.188±0.017 &  & 41.59±0.002 & 0.191±0.001 \\ \hline
\end{tabular}
\end{adjustbox}
\label{tableS4}
\end{table*}

\begin{table*}[htp]
\renewcommand{\thetable}{S\arabic{table}}
\caption{Ablation study on the masking ratios \(\alpha\) (for neighbor nodes) and \(\beta\) (for reference nodes).}
\begin{adjustbox}{width=1\textwidth}
\fontsize{12}{17}\selectfont 
\begin{tabular}{l|lllllllll}
\hline
\multirow{2}{*}{$\alpha$ Value} &  & \multicolumn{2}{c}{Lymph Node} &  & \multicolumn{2}{c}{Pancreas1} &  & \multicolumn{2}{c}{Pancreas2} \\
 &  & \multicolumn{1}{c}{PCC} & \multicolumn{1}{c}{RMSE} &  & \multicolumn{1}{c}{PCC} & \multicolumn{1}{c}{RMSE} &  & \multicolumn{1}{c}{PCC} & \multicolumn{1}{c}{RMSE} \\ \cline{1-4} \cline{6-7} \cline{9-10} 
$\alpha$=0.1 &  & 34.17±0.002 & \textbf{0.220±0.002} &  & 23.20±0.003 & 0.182±0.002 &  & 41.62±0.001 & 0.219±0.003 \\
$\alpha$=0.2 &  & 34.22±0.003 & 0.223±0.002 &  & 23.19±0.004 & 0.180±0.001 &  & 41.33±0.001 & 0.219±0.003 \\
$\alpha$=0.3 &  & 34.04±0.002 & 0.228±0.002 &  & 23.00±0.005 & 0.179±0.001 &  & 41.18±0.002 & 0.217±0.004 \\
$\alpha$=0.4 &  & 34.08±0.002 & 0.227±0.001 &  & 24.82±0.003 & 0.181±0.001 &  & 41.30±0.002 & 0.214±0.003 \\
$\alpha$=0.5 & \textbf{} & 34.19±0.003 & 0.226±0.001 &  & 23.95±0.003 & 0.181±0.002 & \textbf{} & 41.30±0.003 & 0.215±0.003 \\
$\alpha$=0.6 &  & 34.42±0.003 & 0.226±0.002 &  & 24.31±0.005 & 0.179±0.001 &  & 41.03±0.001 & 0.219±0.004 \\
$\alpha$=0.7 &  & 34.73±0.002 & 0.233±0.004 & \textbf{} & 24.01±0.001 & 0.179±0.001 &  & 41.19±0.002 & 0.221±0.001 \\
$\alpha$=0.8 &  & \textbf{35.02±0.003} & 0.225±0.004 & \textbf{} & \textbf{24.48±0.005} & \textbf{0.179±0.003} & \textbf{} & \textbf{41.76±0.003} & \textbf{0.213±0.006} \\
$\alpha$=0.9 &  & 34.70±0.001 & 0.225±0.001 &  & 23.65±0.004 & 0.183±0.004 &  & 41.42±0.001 & 0.222±0.005 \\ \hline
\hline
\multirow{2}{*}{$\beta$ Value} &  & \multicolumn{2}{c}{Lymph Node} &  & \multicolumn{2}{c}{Pancreas1} &  & \multicolumn{2}{c}{Pancreas2} \\
 &  & \multicolumn{1}{c}{PCC} & \multicolumn{1}{c}{RMSE} &  & \multicolumn{1}{c}{PCC} & \multicolumn{1}{c}{RMSE} &  & \multicolumn{1}{c}{PCC} & \multicolumn{1}{c}{RMSE} \\ \cline{1-4} \cline{6-7} \cline{9-10} 
$\beta$=0.1 &  & 34.98±0.002 & 0.227±0.001 &  & 23.19±0.004 & 0.170±0.001 &  & 40.89±0.001 & 0.221±0.002 \\
$\beta$=0.2 &  & 34.73±0.001 & 0.225±0.007 &  & 23.17±0.001 & 0.169±0.001 &  & 40.79±0.002 & 0.222±0.001 \\
$\beta$=0.3 &  & 34.82±0.002 & 0.223±0.003 &  & 23.20±0.003 & 0.170±0.004 &  & 40.70±0.001 & 0.226±0.006 \\
$\beta$=0.4 &  & 34.81±0.001 & 0.228±0.002 &  & 23.41±0.003 & 0.168±0.002 &  & \textbf{41.88±0.002} & \textbf{0.213±0.005} \\
$\beta$=0.5 & \textbf{} & 34.69±0.003 & \textbf{0.221±0.005} &  & 24.29±0.001 & 0.169±0.002 & \textbf{} & 41.66±0.001 & 0.214±0.007 \\
$\beta$=0.6 &  & 34.46±0.001 & 0.223±0.002 &  & 24.28±0.003 & 0.171±0.001 &  & 41.52±0.001 & 0.216±0.003 \\
$\beta$=0.7 &  & 34.33±0.001 & 0.229±0.006 & \textbf{} & 24.43±0.001 & 0.168±0.001 &  & 41.59±0.001 & 0.216±0.002 \\
$\beta$=0.8 &  & 34.91±0.001 & 0.229±0.003 & \textbf{} & 24.43±0.003 & 0.169±0.001 & \textbf{} & 41.47±0.002 & 0.218±0.005 \\
$\beta$=0.9 &  & \textbf{35.02±0.003} & 0.225±0.004 & \textbf{} & \textbf{24.48±0.005} & \textbf{0.179±0.003} & \textbf{} & 41.76±0.003 & 0.213±0.006 \\ \hline
\end{tabular}
\end{adjustbox}
\label{tableS5}
\end{table*}

\newpage
\begin{figure*}[h]
\renewcommand{\thefigure}{S\arabic{figure}}
\centering
\includegraphics[width=1\linewidth]{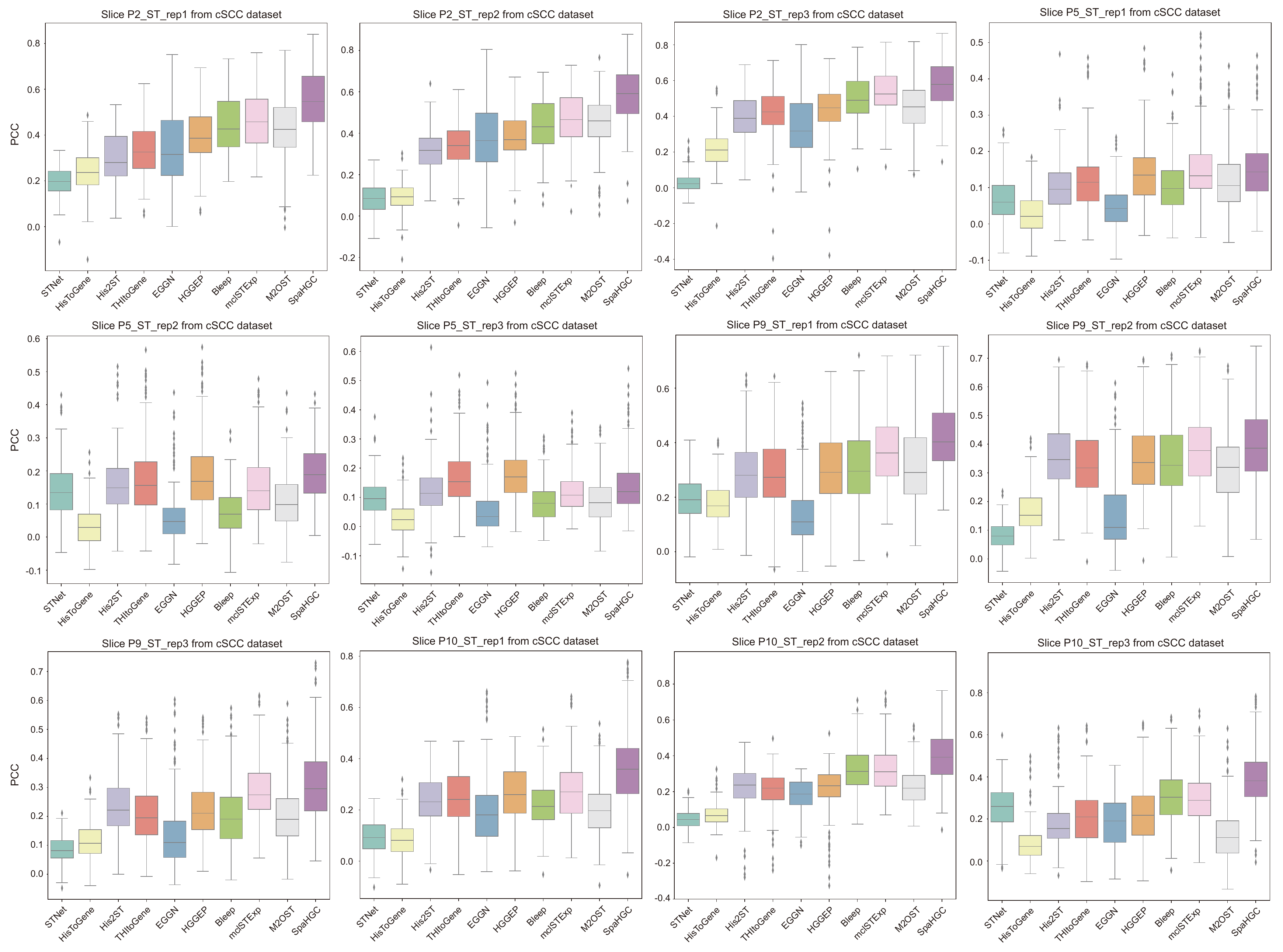}
\caption{Per-slice PCC scores of predicted gene expression on the cSCC datasets.}
\label{figs1}
\end{figure*}

\begin{figure*}[h]
\renewcommand{\thefigure}{S\arabic{figure}}
\centering
\includegraphics[width=1\linewidth]{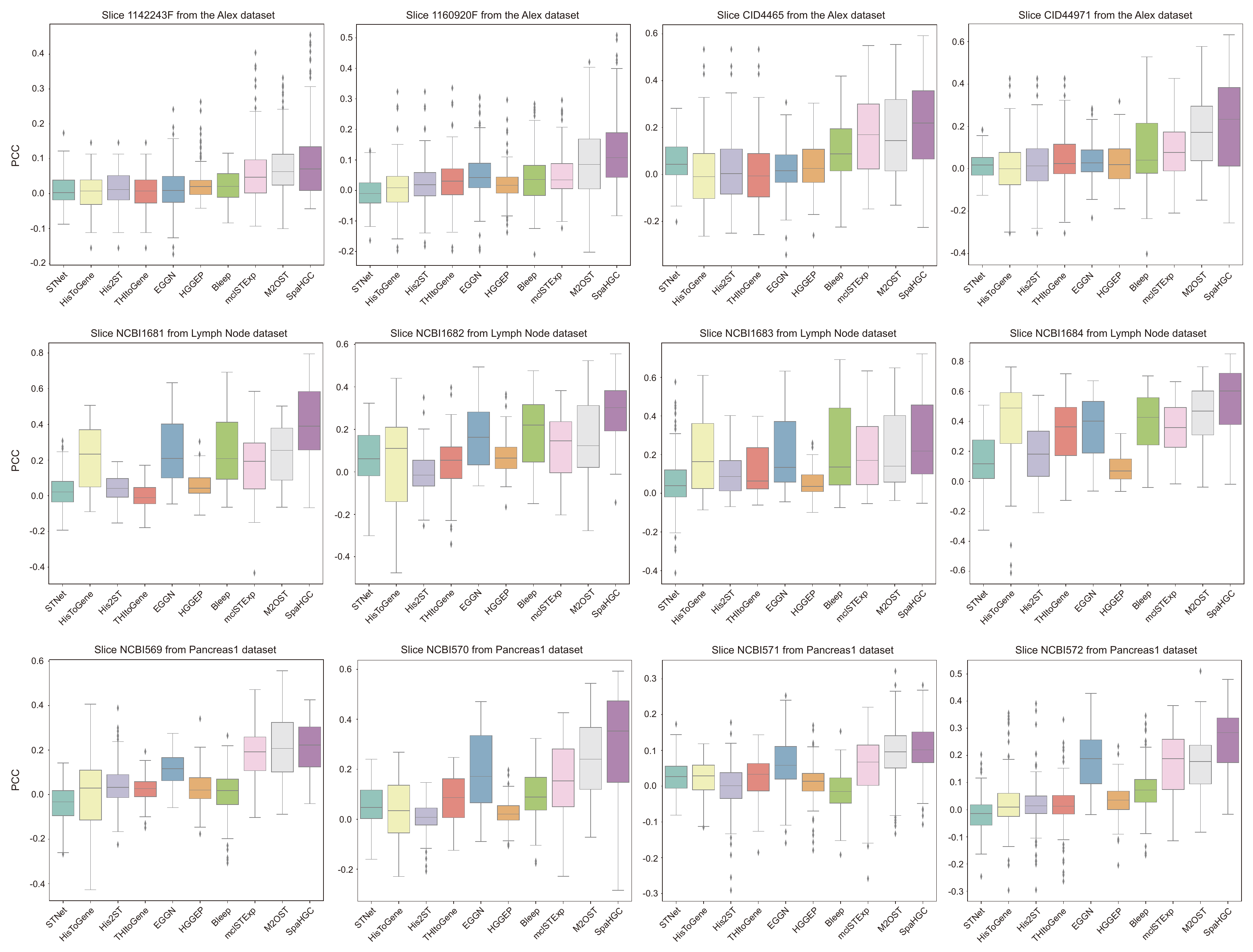}
\caption{Per-slice PCC scores of predicted gene expression across the Alex, Lymph Node, and Pancreas1 datasets.}
\label{figs2}
\end{figure*}

\begin{figure*}[h]
\renewcommand{\thefigure}{S\arabic{figure}}
\centering
\includegraphics[width=1\linewidth]{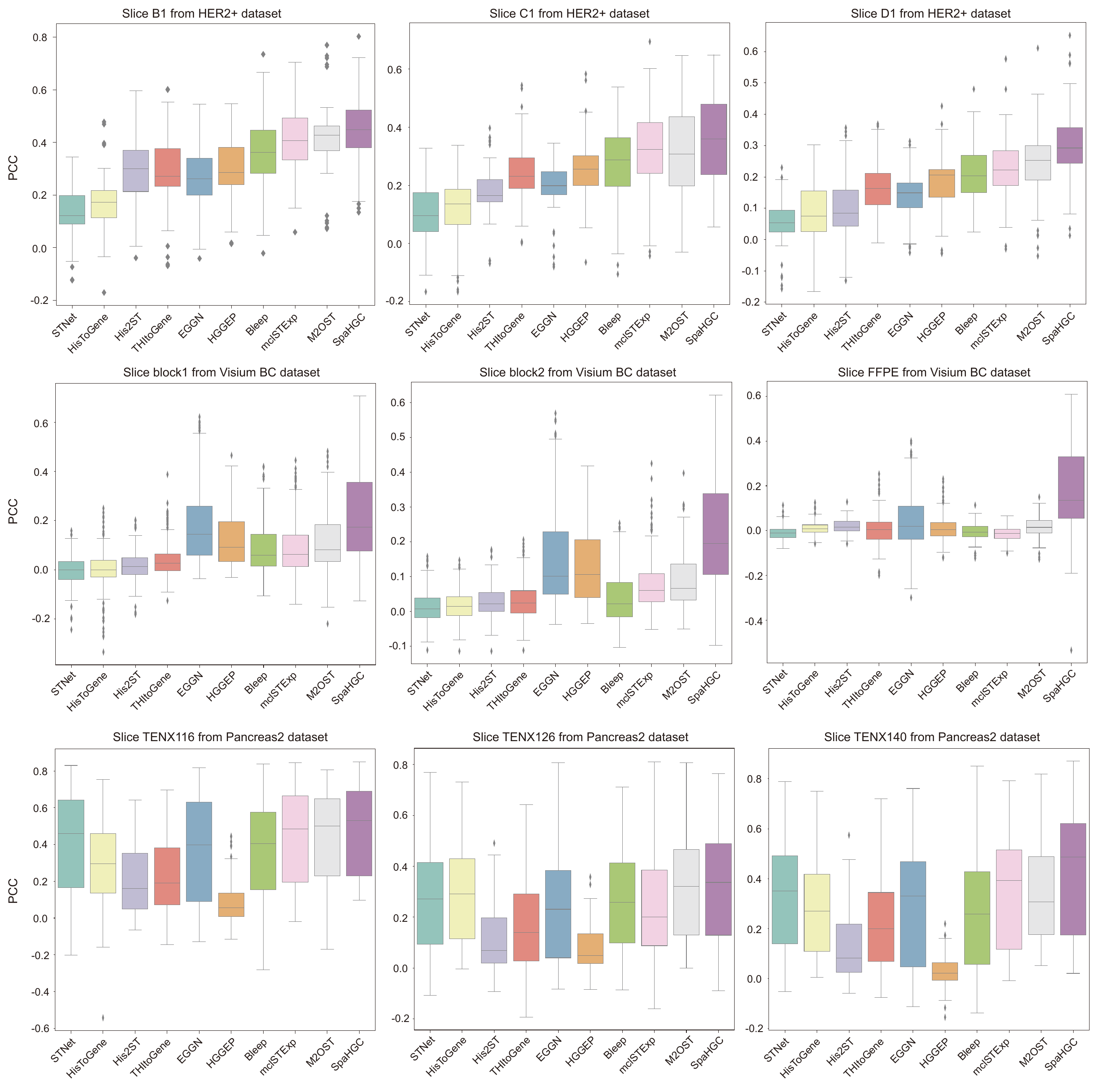}
\caption{Per-slice PCC scores of predicted gene expression across the HER2+, Visium BC, and Pancreas2 datasets.}
\label{figs3}
\end{figure*}

\begin{figure*}[h]
\renewcommand{\thefigure}{S\arabic{figure}}
\centering
\includegraphics[width=1\linewidth]{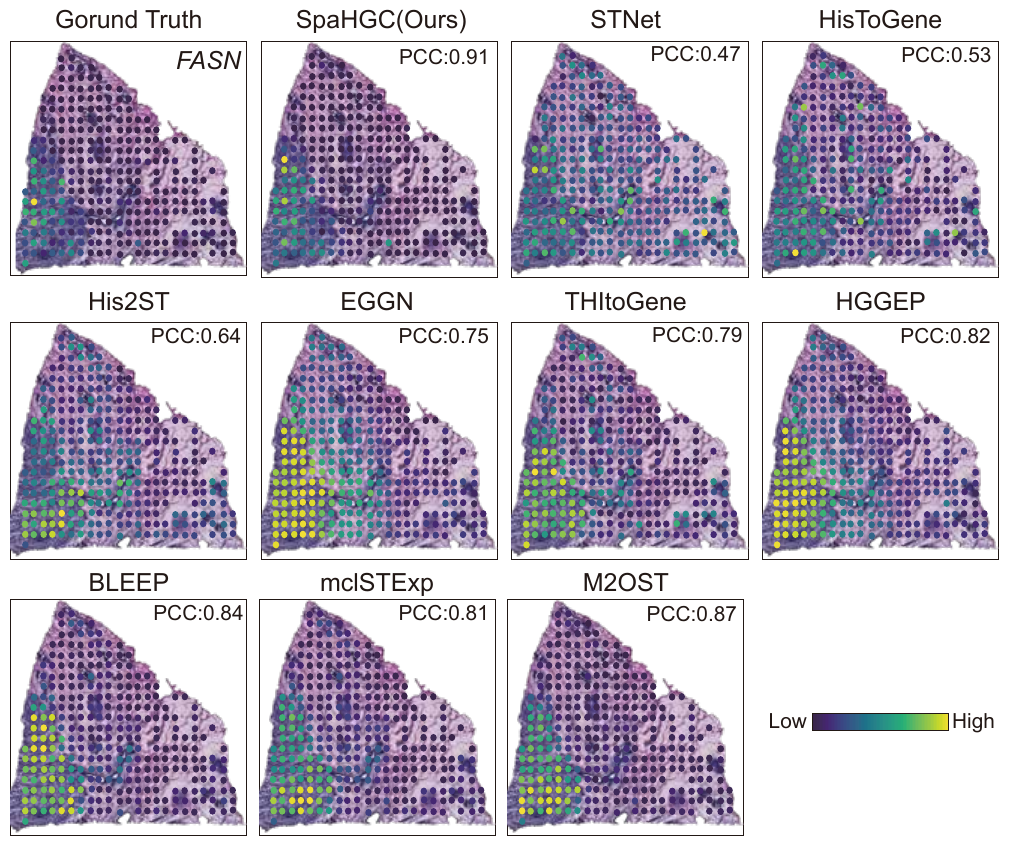}
\caption{Visualization of predicted expression patterns for marker genes \textit{FASN} on the HER2+ dataset.}
\label{figs4}
\end{figure*}

\begin{figure*}[h]
\renewcommand{\thefigure}{S\arabic{figure}}
\centering
\includegraphics[width=0.85\linewidth]{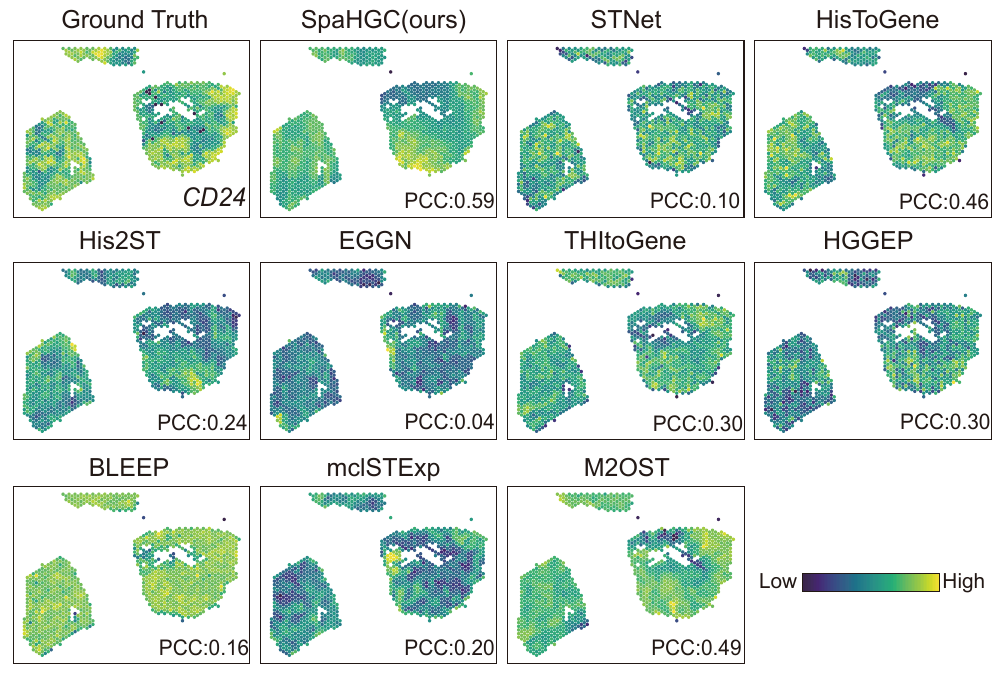}
\caption{Visualization of predicted expression patterns for marker genes \textit{CD24} on the Alex dataset.}
\label{figs5}
\end{figure*}

\begin{figure*}[h]
\renewcommand{\thefigure}{S\arabic{figure}}
\centering
\includegraphics[width=0.85\linewidth]{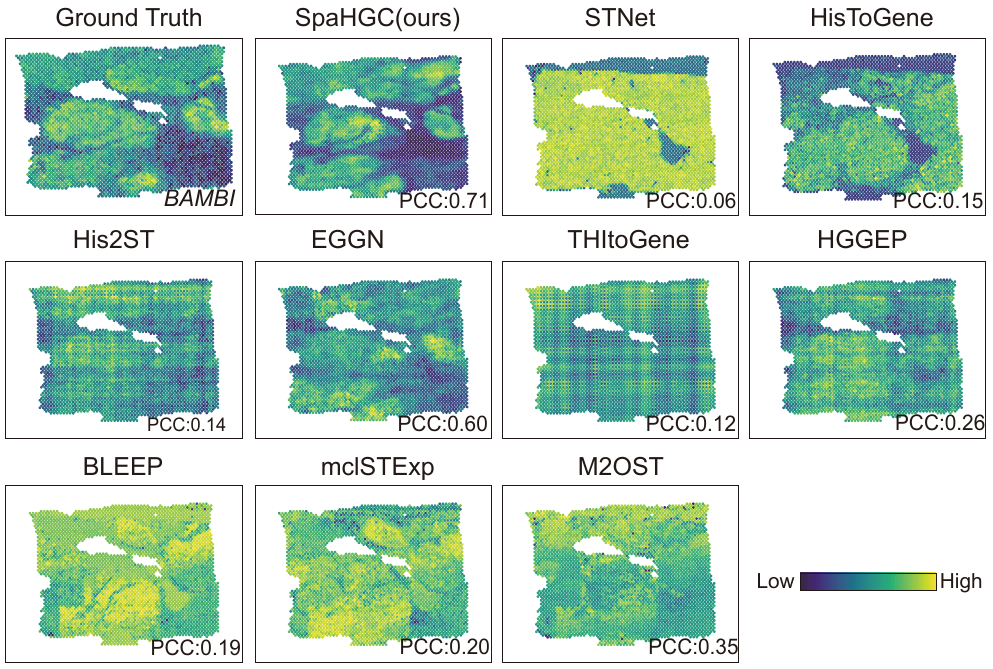}
\caption{Visualization of predicted expression patterns for marker genes \textit{BAMBI} on the Visium BC dataset.}
\label{figs6}
\end{figure*}

\begin{figure*}[h]
\renewcommand{\thefigure}{S\arabic{figure}}
\centering
\includegraphics[width=0.85\linewidth]{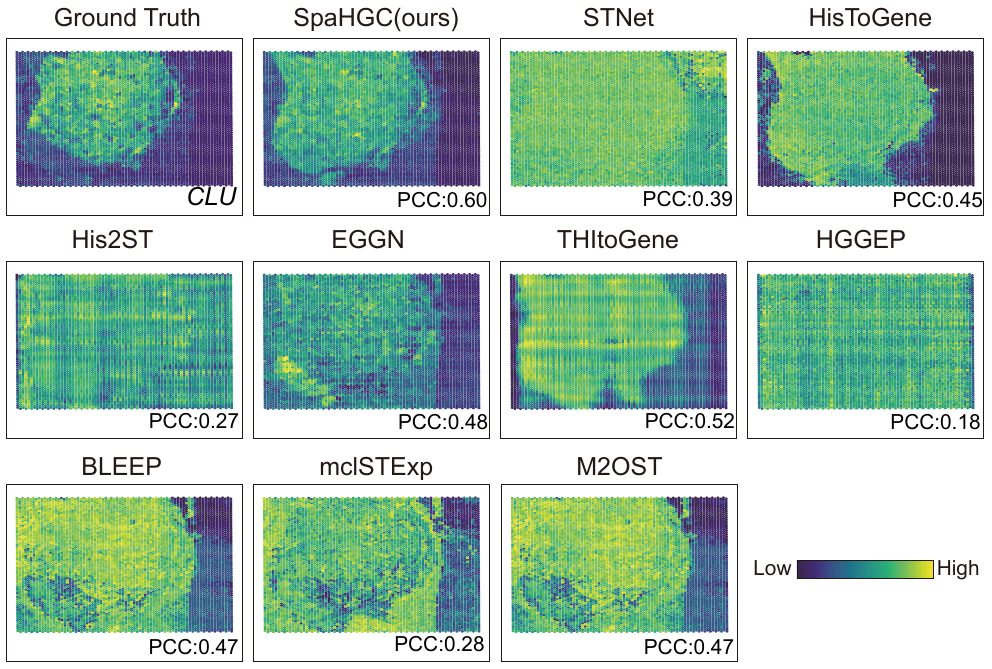}
\caption{Visualization of predicted expression patterns for marker genes \textit{CLU} on the Lymph Node dataset.}
\label{figs7}
\end{figure*}

\begin{figure*}[h]
\renewcommand{\thefigure}{S\arabic{figure}}
\centering
\includegraphics[width=0.85\linewidth]{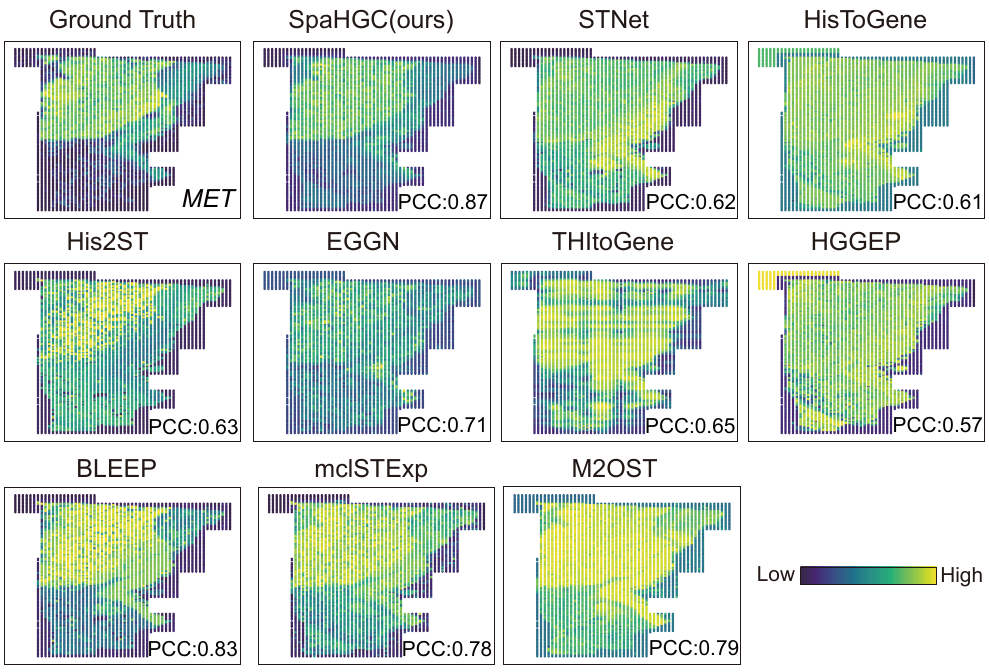}
\caption{Visualization of predicted expression patterns for marker genes \textit{MET} on the Pancreas2 dataset.}
\label{figs8}
\end{figure*}

\begin{figure*}[htp]
\renewcommand{\thefigure}{S\arabic{figure}}
\centering
\includegraphics[width=1\textwidth]{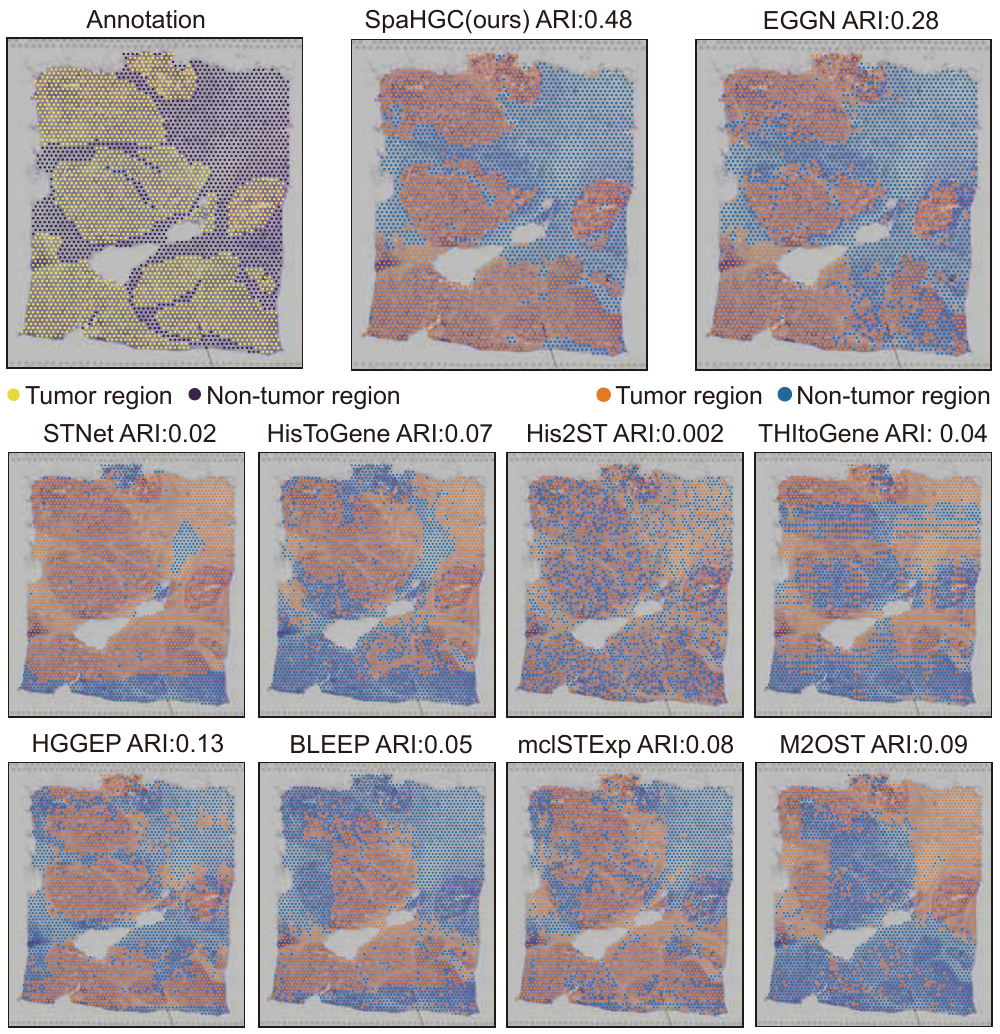}
\caption{Clustering analysis of predicted gene expression profiles on the Visium BC dataset. Spots were grouped into cancerous and normal tissue regions based on pathologist annotations.}
\label{figS9}
\end{figure*}

\begin{figure*}[h]
\renewcommand{\thefigure}{S\arabic{figure}}
\centering
\includegraphics[width=1\linewidth]{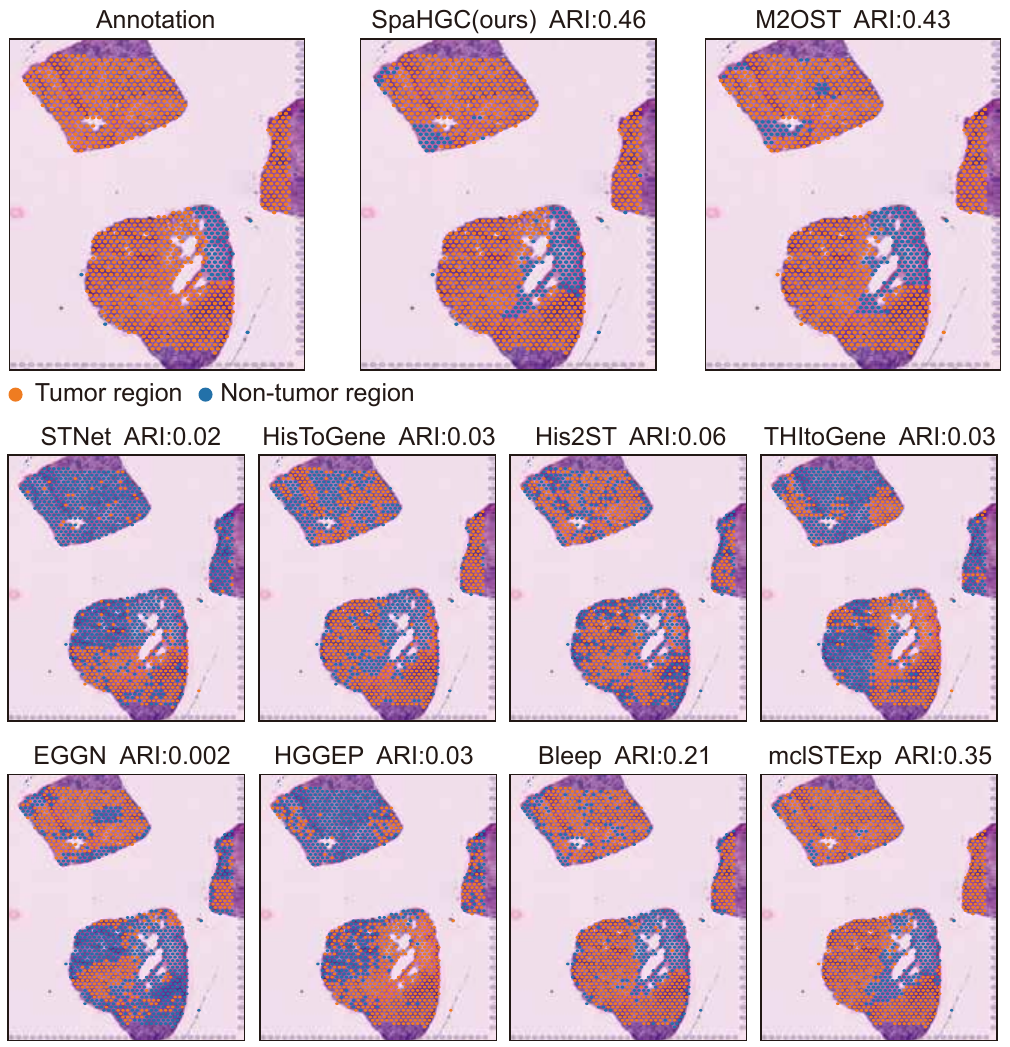}
\caption{Clustering analysis of predicted gene expression profiles on the Alex dataset. Spots were grouped into cancerous and other tissue regions based on pathologist annotations.}
\label{figs10}
\end{figure*}

\begin{figure*}[h]
\renewcommand{\thefigure}{S\arabic{figure}}
\centering
\includegraphics[width=1\linewidth]{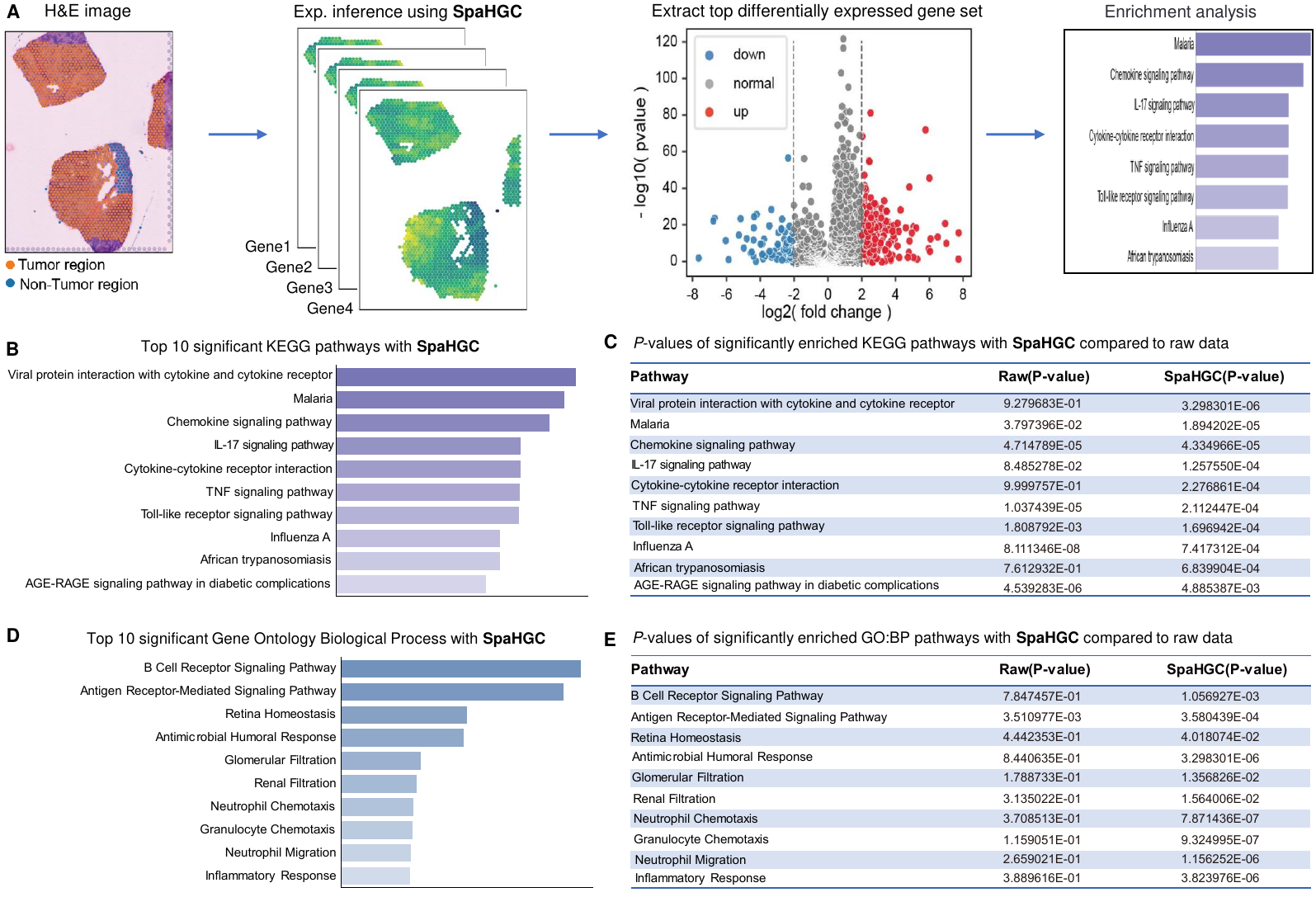}
\caption{
KEGG and GO:BP Pathway Enrichment Analysis of Differentially Expressed Genes in Tumor Regions Compared to Other Tissues, Based on Gene Expression Predicted by SpaHGC on the Alex Dataset.
(\textbf{A}) Workflow diagram for identifying differentially expressed genes and conducting enrichment analysis based on gene expression predictions made by SpaHGC.
(\textbf{B}) and (\textbf{C}) Display the P-values of significantly enriched KEGG pathways identified by SpaHGC compared to the raw dataset.
(\textbf{D}) and (\textbf{E}) Show the P-values of significantly enriched GO:BP pathways identified by SpaHGC compared to the raw data.
}
\label{figs11}
\end{figure*}

\begin{figure*}[t]
\renewcommand{\thefigure}{S\arabic{figure}}
\centering
\includegraphics[width=1\linewidth]{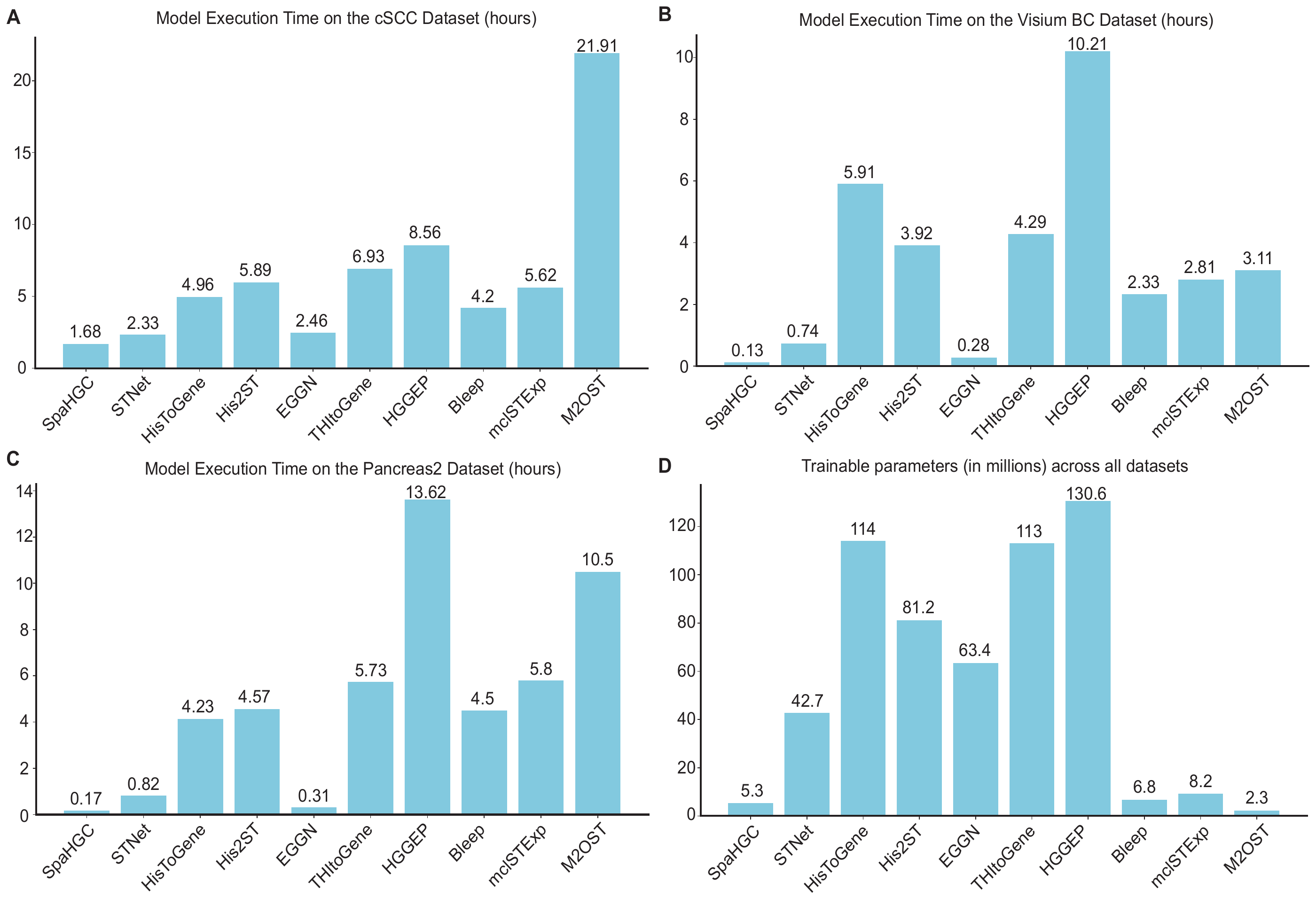}
\caption{Comparison of computational time and trainable parameters between our proposed method and baseline approaches on the same datasets. (\textbf{A}) Runtime on the cSCC dataset (in hours); (\textbf{B}) Runtime on the Pancreas2 dataset (in hours); (\textbf{C}) Inference time on the Pancreas2 dataset (in hours); (\textbf{D}) Number of trainable parameters (in millions) across all datasets.}
\label{figs12}
\end{figure*}

\end{document}